\documentclass{fodsof}
\usepackage{amsmath}
  \usepackage{paralist}
  \usepackage{graphics} 
  \usepackage{epsfig} 
\usepackage{graphicx}  \usepackage{epstopdf}
 \usepackage[colorlinks=true]{hyperref}
\hypersetup{urlcolor=blue, citecolor=red}

\def\currentvolume{X}
 \def\currentissue{X}
  \def\currentyear{200X}
   \def\currentmonth{XX}
    \def\ppages{X--XX}
     \def\DOI{10.3934/fods.2021024}

\usepackage[utf8]{inputenc}
\usepackage{xcolor}
\usepackage{booktabs}
\usepackage{mathtools}
\usepackage{url}
\usepackage{amsmath}
\usepackage{amssymb}
\usepackage{graphicx}
\usepackage{setspace}
\usepackage{pdfpages}
\usepackage{academicons}
\usepackage{xcolor}
\usepackage{adjustbox}

\usepackage{tikz}
\usepackage{autobreak}
\makeatletter

\let\SF@@footnote\footnote
\def\footnote{\ifx\protect\@typeset@protect
    \expandafter\SF@@footnote
  \else
    \expandafter\SF@gobble@opt
  \fi
}
\expandafter\def\csname SF@gobble@opt \endcsname{\@ifnextchar[
  \SF@gobble@twobracket
  \@gobble
}
\edef\SF@gobble@opt{\noexpand\protect
  \expandafter\noexpand\csname SF@gobble@opt \endcsname}
\def\SF@gobble@twobracket[#1]#2{}
\providecommand{\tabularnewline}{\\}

\numberwithin{equation}{section}
\numberwithin{figure}{section}

\usepackage{hyperref}
\usepackage{placeins}
\usepackage{bbold}
\usepackage{subcaption}
\DeclareMathOperator\Hom{Hom}

\DeclareMathOperator\img{img}
\DeclareMathOperator{\Uniform}{Uniform}

\makeatother

\theoremstyle{definition}
\newtheorem{remark}{Remark}

\begin{document}
\sloppy
\title[Generalized Circular Coordinates] 
      {Generalized Penalty for\\ Circular Coordinate Representation}

\author[Hengrui Luo, Alice Patania, Jisu Kim and Mikael Vejdemo-Johansson]{}
\email{hengruiluo@gmail.com}
\email{apatania@iu.edu}
\email{jisu.kim@inria.fr}
\email{mvj@math.csi.cuny.edu}
\thanks{$^*$ Corresponding author: Hengrui Luo}
\maketitle

\centerline{\scshape Hengrui Luo$^{*,1,2}$}
\medskip
{\footnotesize
 \centerline{$^1$Lawrence Berkeley National Laboratory}
   \centerline{1 Cyclotron Rd, Berkeley, CA 94720, USA}
   \smallskip
 \centerline{$^2$Department of Statistics, College of Arts and Sciences}
   \centerline{Ohio State University, 1958 Neil Ave}
   \centerline{Columbus, OH 43210, USA}

} 
\medskip
\centerline{\scshape Alice Patania}
\medskip
{\footnotesize
 \centerline{Indiana University Network Science Institute (IUNI)}
 \centerline{Indiana University}
   \centerline{ 1001 IN-45/46 E SR Bypass}
   \centerline{ Bloomington, IN 47408, USA}
} 
\medskip
\centerline{\scshape Jisu Kim$^{3,4}$}
\medskip
{\footnotesize
 \centerline{$^3$DataShape team, Inria Saclay}
 \centerline{$^4$LMO, Universit{\'e} Paris-Saclay}
   \centerline{B{\^a}timent 307, rue Michel Magat}
   \centerline{Facult{\'e} des Sciences d’Orsay, Universit{\'e} Paris-Saclay}
   \centerline{Orsay, {\^I}le-De-France 91400, France}
} 
\medskip
\centerline{\scshape Mikael Vejdemo-Johansson}
\medskip
{\footnotesize
   \centerline{Department of Mathematics, CUNY College of Staten Island}
   \centerline{Computer Science Programme at CUNY Graduate Center}
   \centerline{2800 Victory Boulevard, 1S-215}
   \centerline{Staten Island, NY 10314, USA}
}

\bigskip

 \centerline{(Communicated by Vasileios Maroulas)}

\begin{abstract}
Topological Data Analysis (TDA) provides novel approaches that allow us to analyze the geometrical shapes and topological structures of a dataset. As one important application, TDA can be used for data visualization and dimension reduction. We follow the framework of circular coordinate representation, which allows us to perform dimension reduction and visualization for high-dimensional datasets on a torus using persistent cohomology. In this paper, we propose a method to adapt the circular coordinate framework to take into account the roughness of circular coordinates in change-point and high-dimensional applications. To do that, we use a generalized penalty function instead of an $L_{2}$ penalty in the traditional circular coordinate algorithm. We provide simulation experiments and real data analyses to support our claim that circular coordinates with generalized penalty will detect the change in high-dimensional datasets under different sampling schemes while preserving the topological structures.
\end{abstract}

\section{\label{sec:Intro}Introduction}

Dimension reduction is one of the central problems in mathematics,
data science and engineering \cite{elad2010sparse,candes2014mathematics}. One of the major challenges in this field is to preserve the topological and geometrical structures of a high-dimensional, nonlinear dataset through dimension reduction.
The nonlinear dimensionality reduction (NLDR) literature \cite{donoho2005new} consists of various attempts to address the problem of representing high-dimensional datasets in terms of low-dimensional coordinate mappings.

Formally, for a dataset $X\subset\mathbb{R}^{d}$ in the form of $X=\{x_{i}=(x_{i,1},x_{i,2},\cdots,x_{i,d})\in\mathbb{R}^{d},\,i=1,\cdots,n\}$
one assumes that $X$ lives on a manifold $M$ and attempts to find a collection of coordinate mappings $\Theta\coloneqq\{\theta_{1},\cdots,\theta_{k}\},\,\theta_{j}:\mathbb{R}^{d}\rightarrow\mathbb{R},\,j=1,\cdots,k$
with $k\leq d$. The reduced dataset can be written as $\Theta(X)=\{(\theta_{1}(x_{i}),\theta_{2}(x_{i}),\cdots,\theta_{k}(x_{i})),$\\$i=1,\cdots,n\}\subset\mathbb{R}^{k}$
through the coordinate mappings. A good choice of coordinate mappings would preserve the main distinctive geometric properties of the manifold.

A well-known dimension reduction method is principal component analysis\\
\noindent (PCA). PCA constructs $k$ linear projections $\theta_{j}$, one for each principal components retained. 
Because of its easily interpretable results, it has become a staple of dimension reduction methods. However, when $M$ has some nontrivial topological structures, these structures may not be preserved by linear dimension reduction methods. Motivated by this, circular coordinates are proposed \cite{de2011persistent}
to take nontrivial topology of $M$ into account when building the coordinate mappings. The paradigm of circular coordinate
representation reveals the intrinsic structure of the high-dimensional
data \cite{wang2011branching}.

The \emph{circular coordinates} are coordinate mappings with circular values in $S^{1}\cong\mathbb{R}/\mathbb{Z}$. The resulting coordinates map
the dataset $X\subset \mathbb{R}^{d}$ to a $k$-torus $\mathbb{T}^{k}=\left(S^{1}\right)^{k}$
through coordinates $\Theta=\{\theta_{1},\cdots,\theta_{k}\},\,\theta_{j}:\mathbb{R}^{d}\rightarrow S^{1},\,j=1,\cdots,k$.
This representation is shown to retain significant topological features while reducing topological noise \cite{de2011persistent}.

The circular coordinates use $L_{2}$ penalty function to ensure that the resulting circular coordinates representation is smooth. However, in many applications such as high-dimensional clustering \cite{de2012minkowski,aggarwal2001surprising}, change point detection \cite{basseville_detection_1993, page_continuous_1954}, etc., it is desirable for the circular coordinates to be similar to a locally constant function and to change more abruptly, so that the pattern of the circular coordinates provides rich information of the topological structure of the dataset. The smooth representation provided by $L_{2}$ penalty is not adequate for this.   Alternatively, the $L_{1}$ penalty function forces the change of the circular coordinates to a small region of the dataset, so that the resulting circular coordinates become more ``locally constant'' and change more abruptly.

In this paper, we propose to impose a generalized penalty for circular coordinates representation, to adjust the roughness of the circular coordinates.
We show by simulations and real data examples that the choice of penalty function could affect the dimension-reduced representations and the detection of topological structures of the dataset.
In particular, we will show how the $L_{1}$ penalty part forces the circular coordinates to be more locally constant and change more abruptly. We will also see how this locally constant pattern can provide rich information such as (quasi-)periodicity or clustering structures.

\subsection{\label{subsec:Circular-coordinate-representati}Circular coordinate representation}\protect\protect

This paper uses several concepts from algebraic topology. In Appendix
\ref{sec:topol-backgr}, we will go through the underlying ideas and
definitions in more detail -- here we will discuss the ``top-level''
ideas with an illustrative example to establish the terminology.

Like standard Topological Data Analysis (TDA) techniques, we approximate the underlying space $M$ by constructing an \emph{approximating complex} $\Sigma$, like \emph{Vietoris-Rips complex} or \emph{\v{C}ech complex} \cite{carlsson2009topology}.  From \cite{de2011persistent}, we can choose an $S^{1}$-valued function on $\Sigma$, known
as the \emph{circular coordinate function}, for each 1-cocycle in $\Sigma$. Intuitively speaking, the circular coordinates are $S^{1}$-valued coordinate functions, which reflect the nontrivial topology of the approximating complex $\Sigma$.
These $S^{1}$-valued functions serve as coordinate maps $\theta$ in the low-dimensional representation.
We use the symbol $\alpha$ to denote a cocycle defined on the underlying complex $\Sigma$.
The pipeline of the circular coordinate representation can be described
as follows:\\
\smallskip
\\
\noindent
\textbf{Procedure of computing circular coordinates with generalized penalty.}
\begin{enumerate}
\item Construct a filtered Vietoris-Rips complex $\Sigma$ to approximate the underlying space where the dataset $X$ lives.
\item Use persistent cohomology as a topological summary to identify the significant 1-cocycles to base the construction of the coordinates upon and discard noise. The significance of the cocycle can be assessed arbitrarily or one can choose every $[\alpha]\in H^{1}(\Sigma,\mathbb{Z}_{p})$ whose persistence
is above a \emph{significant threshold} $\eta$.
\item For each 1-cocycle, we lift the 1-cocycle $[\alpha]$ into
$H^{1}(\Sigma,\mathbb{Z})$ with integer coefficients.
\item\label{step:smooth} For each 1-cocycle, we replace the integer valued cocycle $\alpha$ by a smoothed cohomologous cocycle $\bar{\alpha}$.
\item For each 1-cocycle, we integrate the function $\bar{\alpha}$ to obtain a corresponding $S^{1}$-valued
function $\theta:\Sigma\rightarrow S^{1}$.
\end{enumerate}
This procedure requires the choice of two ``threshold parameters''. The first one is the significance threshold $\eta$ that determines the significant 1-cocycles in step 2. The second one is the distance threshold $r$ that decides which Vietoris-Rips complex to use for computing the smoothed cohomologous cocycle at steps 4 and 5.

\begin{remark}
The choice of significant cocycles defines the circular coordinates. In practice, the 1-cocycle is chosen by looking at their persistence, which may or may not be over a predetermined threshold $\eta$. For simulation experiments, we know the true topology of the dataset and therefore choose the $n$ most persistent 1-cocycles according to the original topology. In our data applications in Sec. \ref{sec:Real-data-analysis} with unknown ground truth, we made a post hoc choice of the significant threshold. How to select the significant cocycles is still an open question in the field: for example, one can separate noise and significant cocycles via statistical inference on persistent homology\cite{fasy2014confidence,michel2015statistical}. (See our implementation at \href{https://github.com/appliedtopology/gcc}{github.com/appliedtopology/gcc}).
\end{remark}

\begin{remark}
The choice of \emph{distance threshold $r>0$} for computing the Vietoris-Rips complex does not influence the outcome of the coordinate process as much as the choice of the 1-cocycle -- as we explore in Appendix~\ref{subsec:distance threshold choice}, the outcome does not change much at all as long as the distance threshold is chosen within the lifetime of the corresponding cocycle. Since the smooth cocycle is optimized on the faces of the  Vietoris-Rips complex, a higher distance threshold impacts mostly on the computational cost of the algorithm: all the optimization steps have computational costs parametrized by the size of the boundary matrix of the complex -- with a smaller complex, the boundary matrix is smaller and the computation is cheaper.
We include a brief discussion of the choice of this threshold in Appendix \ref{subsec:distance threshold choice}.
\end{remark}

\begin{remark}
It is important to stress that when the $H^{1}(\Sigma,\mathbb{Z})$ is trivial, or equivalently there is no significant 1-cocycle in the complex, the circular coordinate methodology cannot be applied, since no nontrivial continuous map $\Sigma\to S^1$ can be found. A trivial $H^1(\Sigma,\mathbb{Z})$ is an \emph{obstruction} to the existence of circular coordinate functions.
\end{remark}

Using the terminology introduced in Appendix \ref{sec:topol-backgr}, we can describe in more detail the theoretical reasoning behind Step~\ref{step:smooth}. The chosen cocycle $\alpha$ can be smoothed to obtain a cohomologous
cocycle $\bar{\alpha}$ that minimizes $L_{2}$ penalty by solving
the following \emph{cohomologous optimization problem} $$\bar{\alpha}=\arg\min\{\|\bar{\alpha}\|_{L_{2}}\mid\exists f\in C^{0}(\Sigma,\mathbb{R}),\bar{\alpha}=\alpha+\delta_{0}f\}.$$
In other words, we are trying to minimize the $L_{2}$ norm of a cocycle (function) $\alpha$ within the collection of cohomologous cocycles (functions) and the resulting $\bar{\alpha}$ can be proven to be harmonically smooth.

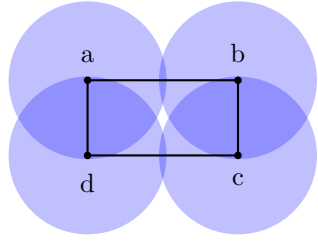
\begin{figure}[t]
\centering

\begin{tikzpicture}
\fill[blue, opacity=0.25] (-1,-0.5) circle (1.05cm);
\fill[blue, opacity=0.25] (1,-0.5) circle (1.05cm);
\fill[blue, opacity=0.25] (-1,0.5) circle (1.05cm);
\fill[blue, opacity=0.25] (1,0.5) circle (1.05cm);
\draw[thick] (-1,-0.5) rectangle (1,0.5);
\fill (-1,-0.5) circle (0.05cm);
\fill (1,-0.5) circle (0.05cm);
\fill (-1,0.5) circle (0.05cm);
\fill (1,0.5) circle (0.05cm);
\node[label={a}] at (-1,0.5) {};
\node[label={b}] at (1,0.5) {};
\node[label=below:{c}] at (1,-0.5) {};
\node[label=below:{d}] at (-1,-0.5) {};
\end{tikzpicture}

\caption{\label{fig:Example-cohomology}Example in Section 2.3 of \cite{zhu2013persistent}
with four points $a=(-1,0.5),b=(1,0.5),c=(1,-0.5),d=(-1,-0.5)$.}
\end{figure}

We illustrate the circular coordinate pipeline using the example in \cite{zhu2013persistent}
as shown in Figure \ref{fig:Example-cohomology}. This example has a chain
complex $0\to C_{1}\to C_{0}\to0$, where the only nonzero boundary
map $\partial$ is given by the matrix
\[
\begin{pmatrix}1 & 0 & 0 & 1\\
-1 & 1 & 0 & 0\\
0 & -1 & 1 & 0\\
0 & 0 & -1 & -1
\end{pmatrix}\Leftrightarrow\begin{array}{ccccc}
 & ab & bc & cd & ad\\
a & 1 & 0 & 0 & 1\\
b & -1 & 1 & 0 & 0\\
c & 0 & -1 & 1 & 0\\
d & 0 & 0 & -1 & -1
\end{array}.
\]
If we choose as a basis for the cochain complex $0\gets C^1\gets C^0\gets0$ the functions $\widetilde{\sigma}(\tau)=\delta_{\sigma\tau}$
given by the Kronecker delta function, then the coboundary map is
simply the transpose of the boundary map, and $\widetilde{a},\widetilde{b},\widetilde{c},\widetilde{d},\widetilde{ab},\widetilde{bc},\widetilde{cd},\widetilde{ad}$
as the basis for the cochain complex as a vector space, where the
tilde denotes the dual element. The coboundary map $\delta$ is given
by the matrix
\[
\begin{pmatrix}1 & -1 & 0 & 0\\
0 & 1 & -1 & 0\\
0 & 0 & 1 & -1\\
1 & 0 & 0 & -1
\end{pmatrix}\Leftrightarrow\begin{array}{ccccc}
 & \widetilde{a} & \widetilde{b} & \widetilde{c} & \widetilde{d}\\
\widetilde{ab} & 1 & -1 & 0 & 0\\
\widetilde{bc} & 0 & 1 & -1 & 0\\
\widetilde{cd} & 0 & 0 & 1 & -1\\
\widetilde{ad} & 1 & 0 & 0 & -1
\end{array}.
\]
Degree 1 cohomology $H^{1}(\Sigma,\mathbb{R})$ is calculated as the quotient of the kernel of
the zero map $C^{1}\to C^{2}=0$ by the image of the coboundary map
$\delta:C^{0}\to C^{1}$. The image of the coboundary map is given
by a basis: $\left\{ \widetilde{ab}+\widetilde{ad},-\widetilde{ab}+\widetilde{bc},-\widetilde{bc}+\widetilde{cd}\right\} $,
and the kernel of the zero map is the entire $C^{1}$. We could complete
\begin{small}
$\left\{ \widetilde{ab}+\widetilde{ad},-\widetilde{ab}+\widetilde{bc},-\widetilde{bc}+\widetilde{cd}\right\} $
\end{small}
into a full basis for $C^{2}$ by adding $\widetilde{ab}$. To construct
an $S^{1}$-valued map from the representative cocycle $\widetilde{ab}$,
for each edge $e$ we would evaluate $\widetilde{ab}(e)$ and take
the resulting number as a \emph{winding number} to ``wrap'' the edge to
the circle. For  $ab$, the resulting winding number is 1 and the rest edges have winding number 0.

We can model a topological circle $S^{1}$ as the quotient space $[0,1]/\langle0\sim1\rangle$,
the unit interval with its endpoints glued together. The circle-valued
map generated by $\widetilde{ab}$ would then send all vertices $a,b,c,d\mapsto0$,
and all points along any of the three other edges would also be sent
to 0. The points along the edge $ab$ would be mapped across the unit
interval: For example, the point $(0.2,0.5)$ on the actual edge $ab$ would be mapped
to a single real value $0.6\in[0,1]$. By using the inclusion map $\mathbb{Z}\to\mathbb{R}$
we can ``lift'' the domain of all of these functions to be real-valued.
Now we can choose an $L_2$ smooth $S^1$-valued map by solving
the cohomologous optimization problem above.

\subsection{Penalty functions}

Circular coordinates are powerful in visualizing and discovering high-dimensional topological structures \cite{wang2011branching}.
The penalty function used in the cohomologous optimization problem is key to the circular coordinate representation of the dataset density.
As a nonlinear dimensionality reduction approach, we want to explore its ability to handle challenges from high-dimensional data analysis.

The circular coordinate is an $S^1$-valued function defined on the dataset and its values could be considered as a representation of the dataset. If there are $k$ circular coordinates, then the represented dataset is on $(S^1)^k$.

The circular coordinate representation relies on the $L_{2}$ penalty function. By changing the penalty function to $L_{1}$ penalty function, the resulting circular coordinate representation becomes more ``locally constant'', that is, the non-locally constant portion of the coordinate function would concentrate near a small region. We will see through simulation studies and real data analyses that adopting $L_{1}$ penalty function is useful in revealing patterns and dynamics.

\subsection{Circular coordinates visualization}
\label{sec:visualize}
\begin{figure}[t!]
\centering

\begin{subfigure}{0.46\linewidth}\centering\includegraphics[height=5.7cm]{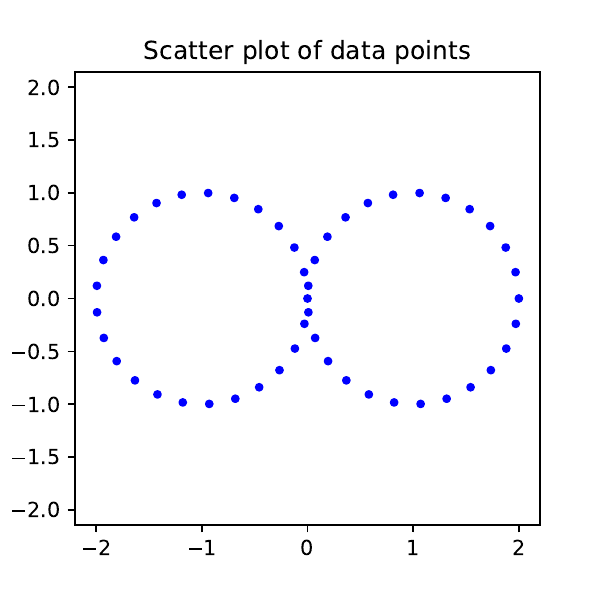}\caption{Scatter plot of $X$.}
\label{fig:cp-cp_dataset}\end{subfigure} \hspace{5mm}  \begin{subfigure}{0.46\linewidth}\centering\includegraphics[height=5.7cm]{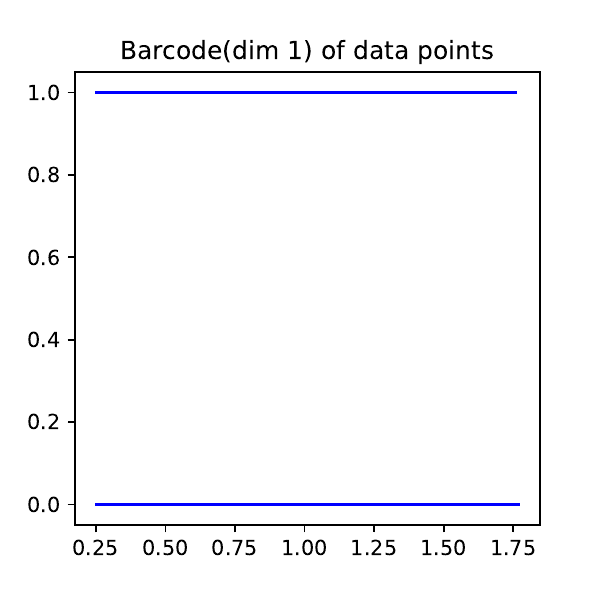}\caption{The $1$-dimensional persistent cohomology of $X$.}
\label{fig:cp-cp_ph}\end{subfigure}

\begin{subfigure}{0.46\linewidth}\centering\includegraphics[height=5.7cm]{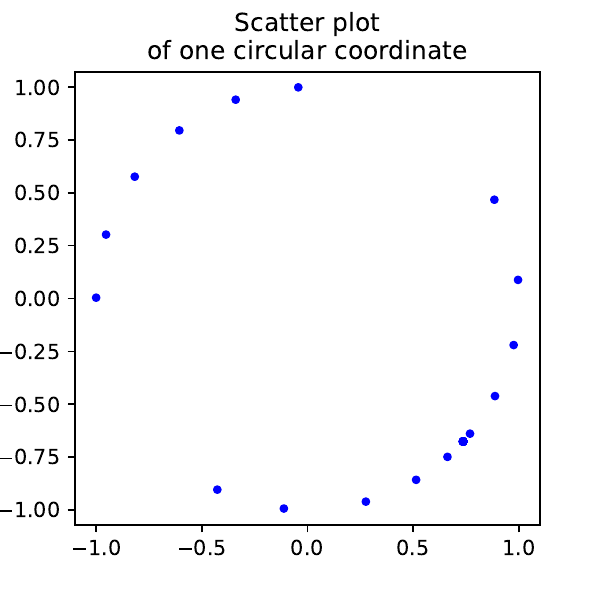}\caption{Scatter plot of one circular coordinate $\theta_{1}(X)$ in $\mathbb{R}^{2}$.}
\label{fig:cp-cp_circle}\end{subfigure} \hspace{5mm}  \begin{subfigure}{0.46\linewidth}\centering\includegraphics[height=5.7cm]{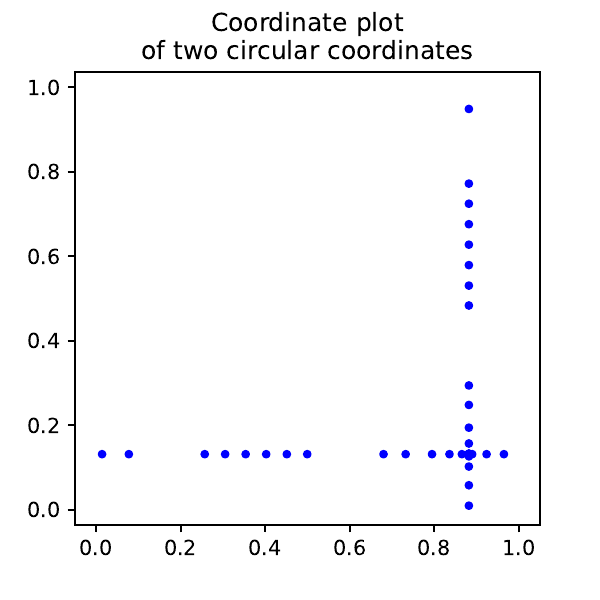}\caption{Coordinate plot of two circular coordinates $\Theta_{12}=(\theta_{1},\theta_{2}):X\to(S^{1})^{2}=\mathbb{T}^{2}$.}
\label{fig:cp-cp_correlation}\end{subfigure}

\begin{subfigure}{0.46\linewidth}\centering
\includegraphics[height=5.7cm,page=1]{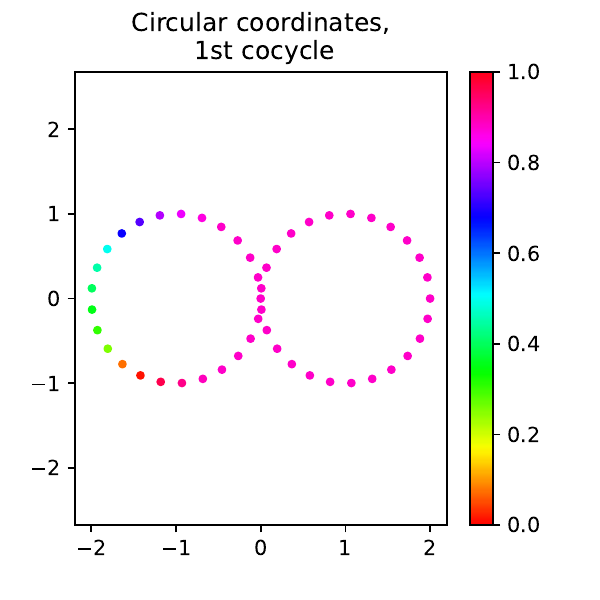}
\caption{Color plot of the first coordinate $\theta_{1}(X)$.}
\label{fig:cp-cp_color1}\end{subfigure}\hspace{5mm} \begin{subfigure}{0.46\linewidth}\centering
\includegraphics[height=5.7cm,page=2]{figs/fig_cp-cp_cc_color}
\caption{Color plot of the second coordinate $\theta_{2}(X)$.}
\label{fig:cp-cp_color2}\end{subfigure}

\caption{The scatter plot, barcode, coordinate plot, and the colormap for the dataset $X\subset\mathbb{R}^{2}$, which is a dataset of $50$ points equidistantly sampled on a figure-$8$ shape. }


\label{fig:cp-cp}
\end{figure}

Circular coordinates can also be used to build a low-dimensional visualization of the  dataset.
We illustrate this with an  example $X\subset\mathbb{R}^{2}$ in Figure \ref{fig:cp-cp}\subref{fig:cp-cp_dataset}.
The $1$-dimensional persistent cohomology of this dataset has two
significant topological features (shown in Figure \ref{fig:cp-cp}\subref{fig:cp-cp_ph}) as two closed loops in $\mathbb{R}^{2}$. Let $\bar{\alpha}_{1},\bar{\alpha}_{2}$
be two significant $1$-cocycles from the persistent cohomology 
and $\theta_{1},\theta_{2}:X\to S^{1}$ be the corresponding circular
coordinates 
under $L_{2}$ penalty.

Since the circular coordinates are values in $S^{1}$, we can
plot the circular coordinates of one such 1-cocycle as points in $\mathbb{R}^{2}$ using a natural embedding map $S^{1}\hookrightarrow\mathbb{R}^{2}$. Figure
\ref{fig:cp-cp}\subref{fig:cp-cp_circle} shows the scatter plot
of one such circular coordinate $\theta_{1}(X)$ in $\mathbb{R}^{2}$.
Although it is straightforward, this representation requires two dimensions to visualize each circular coordinate.

To visualize two (or more) circular coordinates jointly, we use a \emph{coordinate plot}.
The coordinate plot simultaneously visualizes two circular coordinates $\Theta=(\theta_{1},\theta_{2}):X\to(S^{1})^{2}=\mathbb{T}^{2}$
by presenting the $2$-torus $\mathbb{T}^{2}$ as the box $[0,1]^{2}$ but with two horizontal
sides $[0,1]\times\{0\}$ and $[0,1]\times\{1\}$ being identified,
and two vertical sides $\{0\}\times[0,1]$ and $\{1\}\times[0,1]$
being identified. Figure \ref{fig:cp-cp}\subref{fig:cp-cp_correlation}
shows the relation between two circular coordinates $\theta_{1}(X),\theta_{2}(X)$.
The points are lying along a vertical line and a horizontal line
forming a cross, but since the two horizontal sides are identified and
the two vertical sides are identified, the points are indeed lying
on a figure-8 shape on a $2$-torus. Coordinate plots can be extended to visualize more than two circular coordinates, by adding more axes representing individual coordinates.


Alternatively, each circular coordinate can be overlaid on the original dataset
using a color map. Since the circular coordinate values are in $S^{1}$,
it can be translated to a circular color map, such as the cyclic HSV color wheel, to represent mod 1 coordinate values. Figure \ref{fig:cp-cp}\subref{fig:cp-cp_color1} and \subref{fig:cp-cp_color2}
show the color plots of two circular coordinates $\theta_{1}(X)$
and $\theta_{2}(X)$, respectively.

\subsection{Computational complexity and software implementations}

To compute circular coordinates on $n$ datapoints that at the distance threshold $r$ generate $e$ edges and $t$ triangles, we first need to compute the (persistent) cohomology. This can be done in matrix multiplication time~\cite{milosavljevic2011zigzag}. If $M(n)$ is the time complexity of multiplying two $n\times n$-matrices together, then computing (persistent) degree 1 cohomology can be done in $O(M(n)+M(e)+M(t))$. We know that $M(n) = O(n^{2.3755})$~\cite{coppersmithwinograd}, with an asymptotic but impractical current best bound of $M(n) = O(n^{2.3728596})$~\cite{alman2021refined}.

There are several software packages that implement a cohomology-based algorithm and report back the computed cocycles. JavaPlex~\cite{adams2014javaplex} and Dionysus~\cite{dionysus2} are old players in the field, while much improved performance can be found using Ripser~\cite{bauer2019ripser}. The Ripser software package has been reported to comfortably compute with 50\,000 data points, generating over 2\,000\,000 simplices for degree 1 cohomology.

The optimization step also runs in matrix multiplication time -- at least for the $L_2$ case. Using the $L_1$ penalty function we propose in this paper, the optimization step can be pressed down to linear time~\cite{zemel1984n}, leaving the cohomology computation to be the most expensive part of the pipeline.

\subsection{Organization of the paper}

The rest of this paper is organized as follows. We first propose
our method of choosing a generalized penalty function for the circular coordinates and show a specific example
where circular coordinates preserve topological structures in Section
\ref{sec:Cohomologous-optimization-proble}. 
Analyses of simulation
studies in Section \ref{sec:Simulation-studies-under} show how the proposed method behaves under different sampling schemes.
We carry out a real data
analysis with sonar records and congress voting data in Section \ref{sec:Real-data-analysis},
to investigate the performance of our proposed method in real scenarios.
Finally, in Section \ref{sec:Discussion}, we summarize our findings
in this paper and conclude our paper with contributions, discussions, and future work.

\section{\label{sec:Cohomologous-optimization-proble}Generalized penalty
for circular coordinate (GCC) representation}

In this section, we first provide an example where circular coordinates
 may preserve the topology of the dataset while linear dimension reduction
does not, which justifies the use of NLDR methods like circular coordinates. Then we explain how the cohomologous optimization problem that arises in the circular coordinates procedure can be solved with generalized penalty functions, which leads to what we call the \emph{generalized penalty for circular coordinates (GCC)}.

\subsection{\label{sec:PCA-pre-processed-pipeline}Circular coordinates preserves
topology}

\begin{figure}[t]
\centering

\begin{subfigure}{0.45\linewidth}\centering\includegraphics[height=5.7cm]{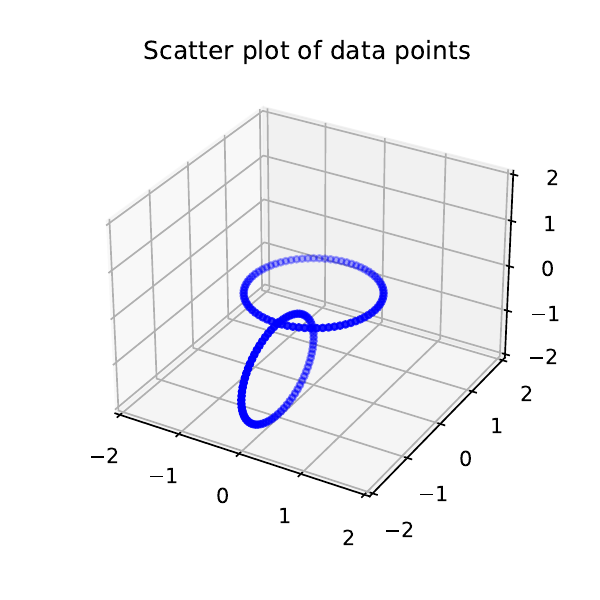}\caption{Scatter plot of $X$.}
\label{fig:cc-pca_dataset_scatter}\end{subfigure} \hspace{5mm} \begin{subfigure}{0.45\linewidth}\centering\includegraphics[height=5.7cm]{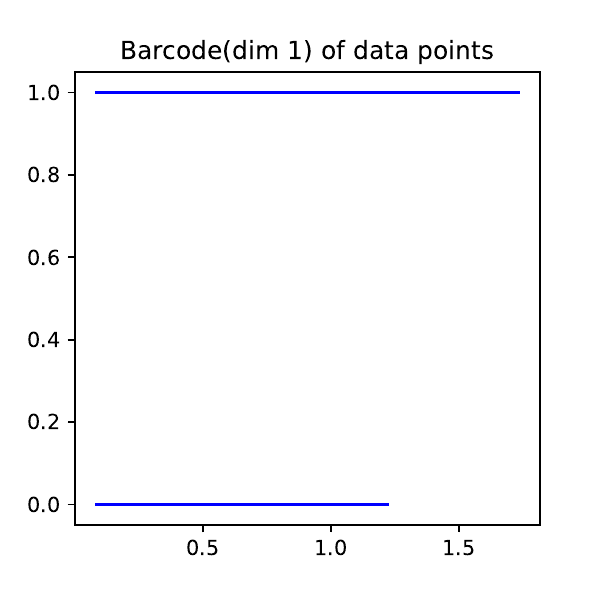}\caption{The $1$-dimensional persistent cohomology of $X$.}
\label{fig:cc-pca_dataset_ph}\end{subfigure}

\caption{The dataset $X\subset\mathbb{R}^{3}$, which is a dataset of $150$ samples
on a figure-$8$ shape $S^{1}\times\{0\}\bigcup\{0\}\times(S^{1}(-1,-1))$, where $S^{1}(-1,-1)$
denotes a unit circle centered at $(-1,-1)$.}

\label{fig:cc-pca_dataset}
\end{figure}

Linear dimension reduction methods, such as PCA, can break down
the topological structure of a dataset when embedding the data into a lower dimension. This loss of information can cause problems when analyzing a dataset. In this section, we show how circular coordinates preserve the topological structure when a linear dimension reduction method (i.e., PCA) fails to do so. To show this, we created a dataset $X\subset\mathbb{R}^{3}$ in Figure \ref{fig:cc-pca_dataset}\subref{fig:cc-pca_dataset_scatter},
formed by  points equidistantly sampled from two circles in $\mathbb{R}^3$ touching  orthogonally at one point.
As expected, the $1$-dimensional persistent cohomology of this dataset shows two significant topological features as  (shown in Figure \ref{fig:cc-pca_dataset}\subref{fig:cc-pca_dataset_ph}) two closed loops in $\mathbb{R}^{3}$.

\begin{figure}[t]
\centering

\begin{subfigure}{0.45\linewidth}\centering\includegraphics[height=5.7cm]{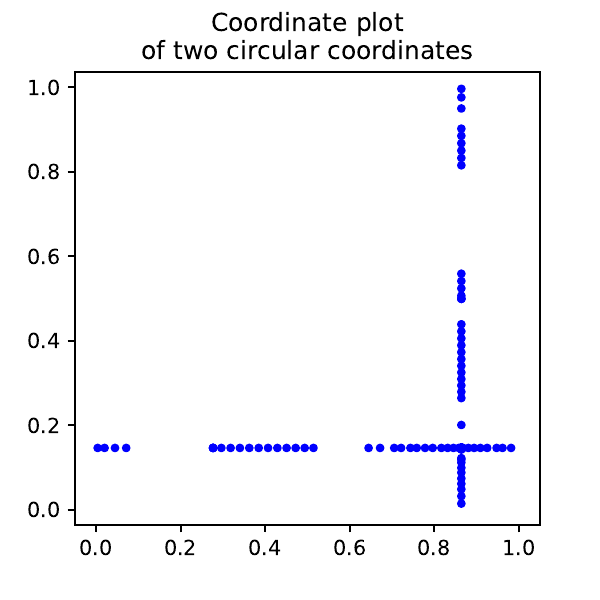}\caption{Coordinate plot of $X$}
\label{fig:cc-pca_cc_correlation}\end{subfigure} \hspace{5mm} \begin{subfigure}{0.45\linewidth}\centering\includegraphics[height=5.7cm]{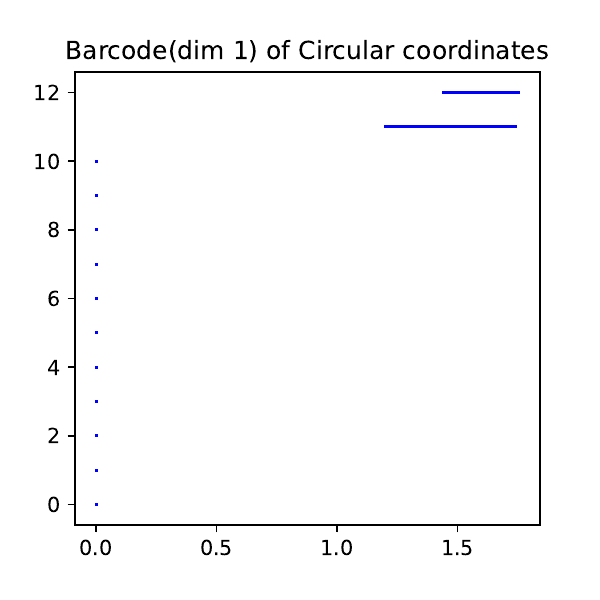}\caption{The $1$-dimensional persistent cohomology of $X^{cc}$.}
\label{fig:cc-pca_cc_ph}\end{subfigure}\caption{The dimension reduced data $X^{cc}$ obtained from circular coordinates based on  the Vietoris-Rips complex constructed from $X$.}

\label{fig:cc-pca_cc}
\end{figure}

 On one hand, we can compute the circular coordinate representation. Let $\bar{\alpha}_{1},\bar{\alpha}_{2}$
be two significant $1$-cocycles chosen from the persistent cohomology of
the Vietoris-Rips complex constructed from $X$ in Figure \ref{fig:cc-pca_dataset}\subref{fig:cc-pca_dataset_ph},
and let $\theta_{1},\theta_{2}:X\to S^{1}$ be the corresponding circular
coordinates for the cocycles $\bar{\alpha}_{1},\bar{\alpha}_{2}$
under $L_{2}$ penalty $\|\bar{\alpha}\|_{L_{2}}$. Let $\Theta=(\theta_{1},\theta_{2}):X\to(S^{1})^{2}=\mathbb{T}^{2}$,
and we obtain the dimension reduced data as $X^{cc}:=\Theta(X)\subset \mathbb{T}^{2}$
embedding in a $2$-torus.  The coordinate plot of $X$, which is also the scatter plot of $X^{cc}$ on the torus, is shown in \ref{fig:cc-pca_cc}\subref{fig:cc-pca_cc_correlation}. We see that the
persistent cohomology of $X^{cc}$ in Figure \ref{fig:cc-pca_cc}\subref{fig:cc-pca_cc_ph}
contains two persistent 1-cocycles, as for the original dataset
$X$ in Figure \ref{fig:cc-pca_dataset}\subref{fig:cc-pca_dataset_ph}.


\begin{figure}[t]
\centering

\begin{subfigure}{0.45\linewidth}\centering\includegraphics[height=5.7cm]{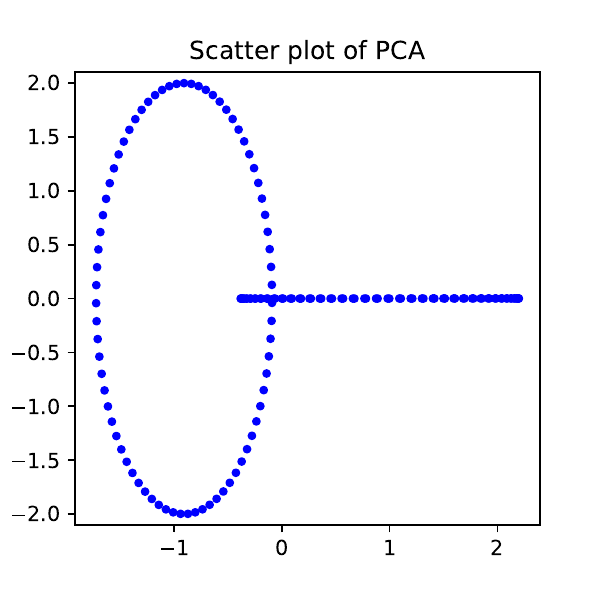}\caption{Scatter plot of $X^{pca}$.}
\label{fig:cc-pca_pca_scatter}\end{subfigure} \hspace{5mm} \begin{subfigure}{0.45\linewidth}\centering\includegraphics[height=5.7cm]{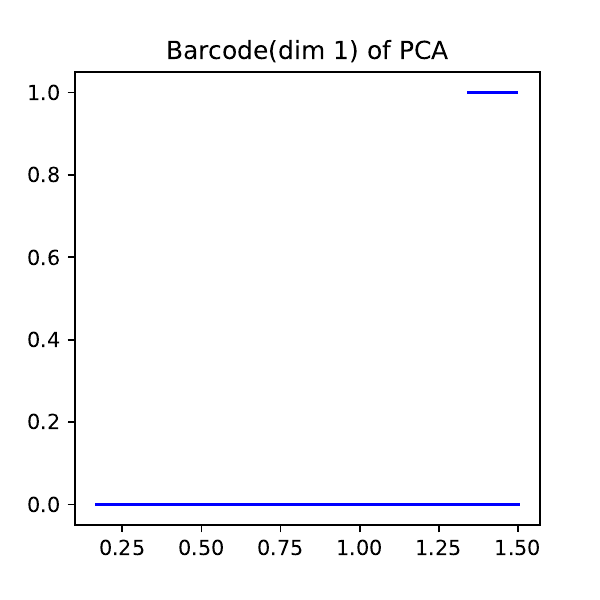}\caption{The $1$-dimensional persistent cohomology of $X^{pca}$.}
\label{fig:cc-pca_pca_ph}\end{subfigure}

\caption{The PCA representation $X^{pca}$ from choosing $2$ principal components.}

\label{fig:cc-pca_pca}
\end{figure}

On the other hand, we choose to consider the embedding defined
by the first 2 principal components for comparison. From Figure \ref{fig:cc-pca_pca}\subref{fig:cc-pca_pca_scatter},
we can see that one of the 1-dimensional coboundaries of the original data $X$ collapsed and the $1$-dimensional cohomology structure of $X$ is distorted in $X^{pca}$. And the collapsed $1$-dimensional cohomology
structure is also unidentifiable
using persistent cohomology of the embedded dataset $X^{pca}$ as seen in Figure~\ref{fig:cc-pca_pca}\subref{fig:cc-pca_pca_ph}. In Appendix  \ref{sec:PCA-pre-processed-pipeline_gpca}, we also analyze with the generalized PCA (GPCA) representation \cite{VidalMS2005}, and find that
the topological structure still gets distorted in the lower dimensional representation.

The example above shows that circular coordinate representations preserve important topological and geometrical structures in the dataset, which are easily distorted by linear dimension reduction methods. We will see more simulations
about how circular coordinates can effectively preserve the topological structures in Section \ref{sec:Simulation-studies-under}.
\FloatBarrier
\subsection{Cohomologous optimization problem with generalized penalty functions}

As we previously discussed, circular coordinates can be obtained by
solving the following \emph{cohomologous optimization problem}:
\begin{equation}
\bar{f}=\arg\min_f\{\|\bar{\alpha}\|_{L_{2}}\mid f\in C^{0}(\Sigma,\mathbb{R}),\bar{\alpha}=\alpha+\delta_{0}f\}.\label{eq:cohomologous opt - L2}
\end{equation}
When using the $L_{2}$ penalty, \cite{de2011persistent} proved
that the constructed coordinates possess harmonic smoothness and other
well-behaved properties. Usually, on a low-dimensional dataset with
significant topological features, this $L_{2}$ penalty works well
and detects features by showing changes in coordinate values (as shown in Figure \ref{fig:cp-cp} (e)(f)).


We propose to use a generalized penalty function in the optimization problem (\ref{eq:cohomologous opt - L2}) instead of the usual $L_2$ penalty.

The circular coordinate for a
1-cocycle $\alpha$ will be the solution of the following optimization
problem:
\begin{equation}
\bar{f}=\arg\min_f\{(1-\lambda)\|\bar{\alpha}\|_{L_{1}}+\lambda\|\bar{\alpha}\|_{L_{2}}\mid f\in C^{0}(\Sigma,\mathbb{R}),\bar{\alpha}=\alpha+\delta_{0}f\}.\label{eq:elastic net}
\end{equation}
In particular, when $\lambda =1$, the penalty reduces to $L_2$ penalty. When $\lambda=0$, we have the following form using
only an $L_{1}$ penalty function,
\begin{equation}
\bar{f}=\arg\min_f\{\|\bar{\alpha}\|_{L_{1}}\mid f\in C^{0}(\Sigma,\mathbb{R}),\bar{\alpha}=\alpha+\delta_{0}f\}.\label{eq:cohomologous opt - L1}
\end{equation}
For a finite dataset $X=\{x_{1},\cdots,x_{n}\}$, each $\mathbb{R}$-function
$f$ can be represented as an $n$-vector $x_{f}\coloneqq\left(f(x_{1}),\cdots,f(x_{n})\right)\in\mathbb{R}^{d}$.
Note that these two problems (\ref{eq:cohomologous opt - L2}) and
(\ref{eq:cohomologous opt - L1}) above, can be formalized as a restrained
optimization problem, since coboundary maps $\delta$ are linear operators
by definition.

With the $L_1$ penalty part in \eqref{eq:elastic net}, the circular coordinates function becomes more ``locally constant'' and changes more abruptly. That is, the non-locally constant portion of the coordinate functions would be forced to concentrate near a small region. 
Hence, the class of possible coordinate functions is restricted to nearly locally constant functions. When viewed as being embedded in the circle $S^{1}$, the circular coordinates become more sparsely distributed in $S^{1}$. This provides an alternative circular coordinate representation than the harmonic smooth representation given by $L_2$ penalty. And from \eqref{eq:elastic net}, we can balance between the nearly locally constant representation coming from $L_{1}$ penalty and the smooth representation coming from $L_2$ penalty. We will show
by simulation studies and real data analyses that this representation still preserves the topological features and
provides rich information of the topological structure of the dataset
in  practice. This idea motivates our formulation and  explorations in this paper.

\section{\label{sec:Simulation-studies-under}Simulation studies}

In this section, we provide simulation studies and examine
their circular coordinates embeddings under both $L_{2}$ and
a generalized penalty function.

As remarked in Section \ref{subsec:Circular-coordinate-representati}, before we compute the coordinates, we need to choose a significant
cocycle(s) from the (Vietoris-Rips) persistent cohomology based on the dataset. 
The existence of a 1-cocycle in the dataset indicates that there exists a nontrivial continuous
map to $S^{1}$. 
If the dataset does
not possess significant geometric structures, circular coordinates
may not be the right method to use.
In the following simulation studies,
our datasets are sampled from manifolds with known topology, for which we know there are significant cocycles. We choose cocycles that are consistent with the ground-truth topology of each example to compute the generalized circular coordinates.

To highlight the smoothness of the coordinate function 
induced by the choice of the penalty function, we highlight the  edges in the complex over which the coordinate function does not change its value. We will denote these edges as  \emph{constant edges}.
In our simulation study, we choose to visualize 
as constant those
edges of the Vietoris-Rips complex over which the value change of the coordinate functions does not exceed $10^{-5}$, which is used by default in our implementation.
We display the circular coordinate values (mod 1) using the color scale described in Section \ref{sec:visualize}.

To verify the validity of the computed circular coordinates with the topological ground truth, we study the relationship between the circular coordinates (Y-axis)
and the angle formed by each data point (X-axis, computed as $\text{arc tan}(x_{2}/x_{1})$
for $x=(x_{1},x_{2})\in\mathbb{R}^{2}$) using (angle) \emph{correlation plots}. The range of x-axis of correlation plot is usually set to be $[0,2\pi)$; the range of y-axis of correlation plot is usually set to be $[0,1]$.

In Section \ref{subsec:Parameter-uniform-sampling-scheme}, we will observe
how the circular coordinates behave under $L_{2}$ and generalized
penalty functions. We will show that, as a result, different types of penalty functions lead to significant differences in the final representations. \cite{robinson2019} has
suggested that  coordinates with generalized penalty might detect geometric features
such as strata in a stratified manifold setting. We investigate this
conjecture computationally using the roughness introduced by a generalized
penalty
in Section \ref{subsec:Example-3:-Dupin}.

Based on our experiments and a personal communication \cite{holmes2020}, we know that
the sampling scheme usually affects the sparsity of the dataset, and the results obtained from TDA method. To investigate this further, 
in Appendix \ref{subsec:Volume-uniform-sampling-scheme}, we compare and discuss the results obtained in the simulation studies under parameter uniform and volume uniform sampling schemes.

For all simulations, to solve the cohomologous optimization problem (\ref{eq:cohomologous opt - L2}) and (\ref{eq:cohomologous opt - L1}), we use Adams optimizer \cite{kingma2014adam}
with learning rate parameter $\epsilon=10^{-4}$ and 1000 steps of
iterations. When solving the cohomologous optimization problem
with generalized penalty functions, the numerical issues become more
subtle compared to the $L_{2}$ case, which can usually be solved by choosing appropriate learning rates.

\subsection{\label{subsec:Parameter-uniform-sampling-scheme}Simulation studies}

We sample data points from known parametrized manifolds using a \emph{parameter uniform sampling scheme}. For example,
on a two-dimension-al disc $\{(x,y)\in\mathbb{R}^{2}\mid x^{2}+y^{2}\leq1\}$,
we may have a parameterization $(r,\theta)\mapsto(r\cos\theta,r\sin\theta)\in\mathbb{R}^{2}$.
Then a parameter uniform sampling scheme subject to this parameterization means
that we sample the parameters uniformly at random from $r\sim\Uniform(0,1)$, $\theta\sim\Uniform(0,2\pi)$.

In other words, we sample the points lying on the chosen manifold with a uniform
sampling scheme on the parameter space of a specific parameterization of the manifold.

\subsection{\label{subsec:Example-1:-Ring}Example 1: Ring }

\begin{figure}[t!]
\centering

\begin{tabular}{cc}
circular coordinates & correlation plot\tabularnewline
\hline
\includegraphics[height=5.7cm,page=10]{figs/Example1_d=1.5_output_edit.pdf} & \includegraphics[height=5.7cm]{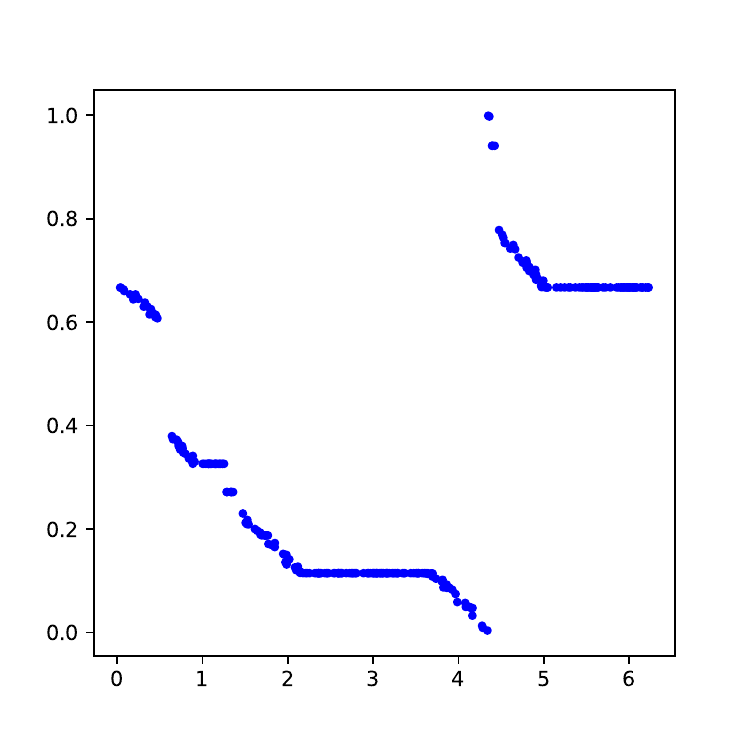} \tabularnewline
\includegraphics[height=5.7cm,page=17]{figs/Example1_d=1.5_output_edit.pdf}  & \includegraphics[height=5.7cm]{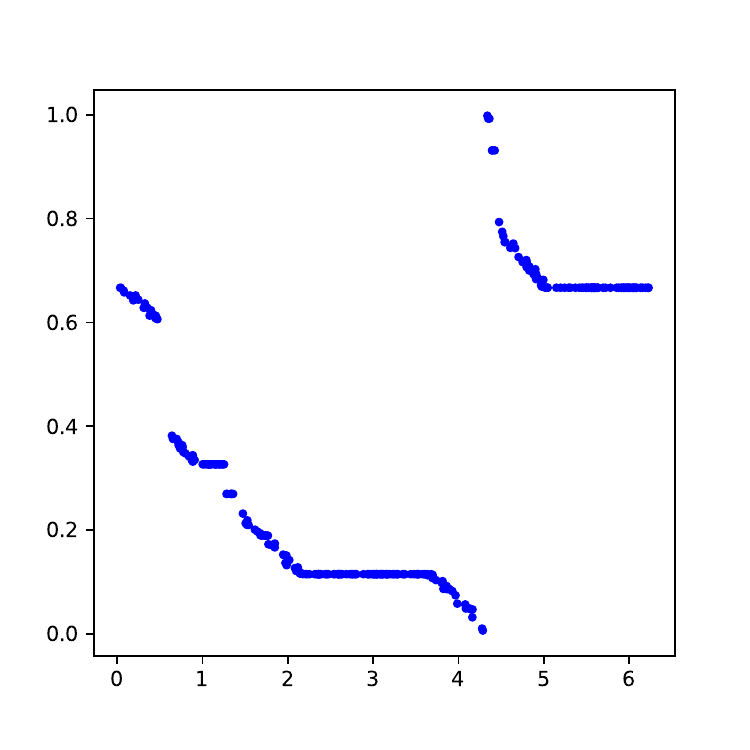}  \tabularnewline
\includegraphics[height=5.7cm,page=24]{figs/Example1_d=1.5_output_edit.pdf}  & \includegraphics[height=5.7cm]{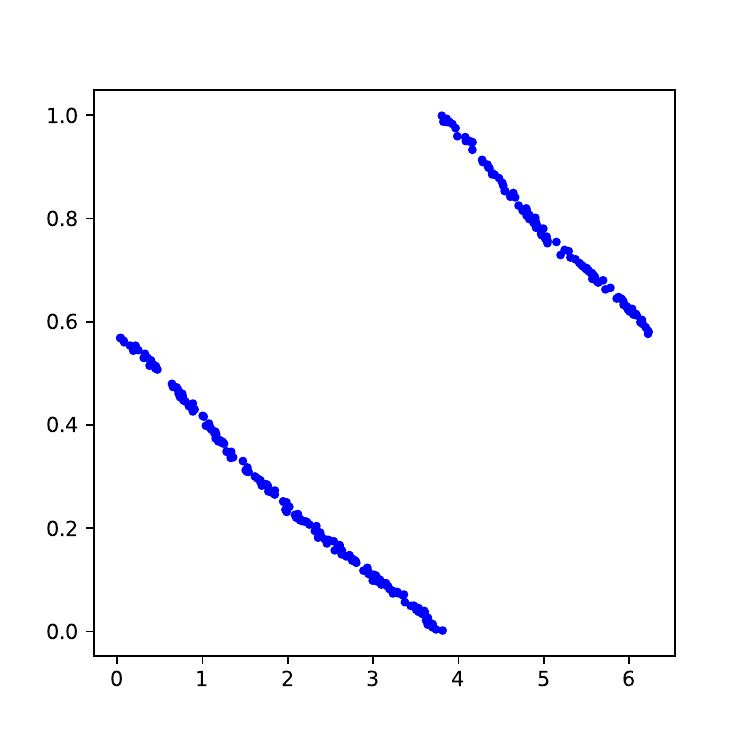} \tabularnewline
\end{tabular}

\caption{\label{fig:Example-1:-Ring}Example 1: The $L_{2}$ smoothed and generalized
penalized circular coordinates of the uniformly sampled dataset ($n=300$)
from a ring of inner radius $R=1.5$ and width $d=1.5$. The first,
second, and the third row correspond to $\lambda=0,0.5$, and 1, respectively.}
\end{figure}

For our first example, we consider a ring (a.k.a. annulus) in $\mathbb{R}^2$, with width $d=1.5$ and inner radius $R=1.5$.
We sample 300 points $(r,\theta)$ at random in the square $[0,1]\times[0,2\pi]$,
and then we use a standard parameterization $(r,\theta)\mapsto((R+rd)\cos\theta,(R+rd)\sin\theta)$
to map the points $(r,\theta)$ onto an area in $\mathbb{R}^{2}$
as shown in Figure \ref{fig:Example-1:-Ring}. The example manifold has only one connected component and a single 1-cocycle. We choose the most
significant cocycle and obtain circular coordinates for this dataset. We can see the concentration of constant edges in circular coordinates computed with the generalized penalty function, and fewer ones in the $L_2$ case.

In this experiment, we also studied the sensitivity of generalized circular coordinates with respect to the choice of coefficient $\lambda$ as defined in equation (\ref{eq:cohomologous opt - L1}). The primary relevant problem in applications of circular coordinates is how the choice of $\lambda$ could affect the norm of the resulting (optimized) coordinate functions. In Figure \ref{fig:Example-sensitivity}, we displayed the  $L_1$, $L_2$, and the mixed norm $(1-\lambda)\|\bar{\alpha}\|_{L_{1}}+\lambda\|\bar{\alpha}\|_{L_{2}}$ of the coordinate function as the coefficient $\lambda$ changes. It is not hard to see that the sharp difference of coordinates in terms of $\lambda$ happens when $\lambda$ is close to 1. The $L_1$ and $L_2$ norms present a drop near $\lambda=1$ but the mixed norm show monotonicity except for numerical round-off near $\lambda=1$.  Empirical observations from this set of experiments lead us to believe that it suffices to consider only situations where $\lambda=0,0.5,1$ in the subsequent discussions.
As can be seen in the vertical scales of Figure~\ref{fig:Example-sensitivity}, the $L_1$ and $L_2$ norms differ by an order of magnitude (just around 150 for the $L_1$ norm in this example, just around 10 for the $L_2$ norm), so that for most choices of $\lambda$, the mixed norm penalty is dominated by the $L_1$ norm term.

\begin{figure}[t!]
\centering

\begin{tabular}{cccc}
$\lambda=0$ & $\lambda=1$ &\tabularnewline
\midrule
\includegraphics[height=5.7cm,page=10]{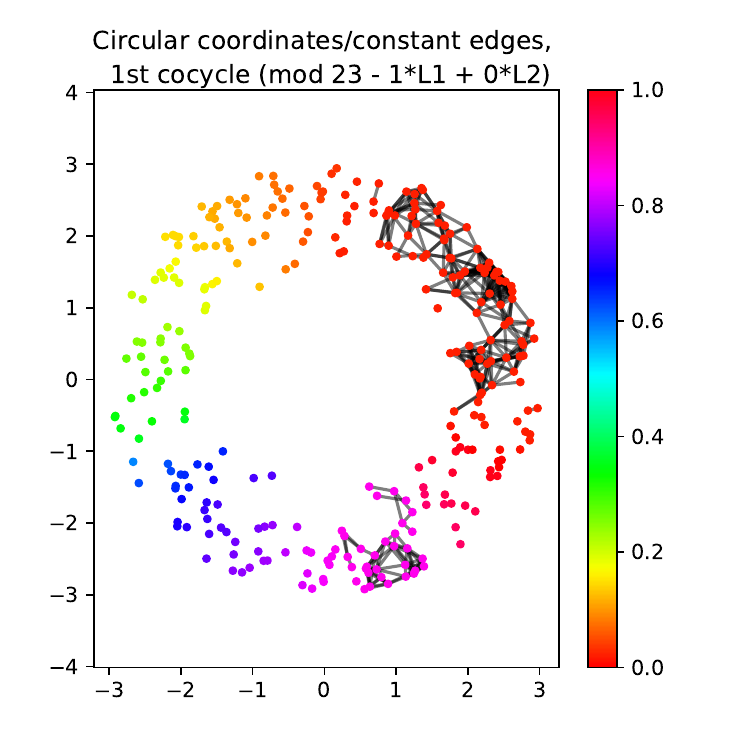}  & \includegraphics[height=5.7cm,page=24]{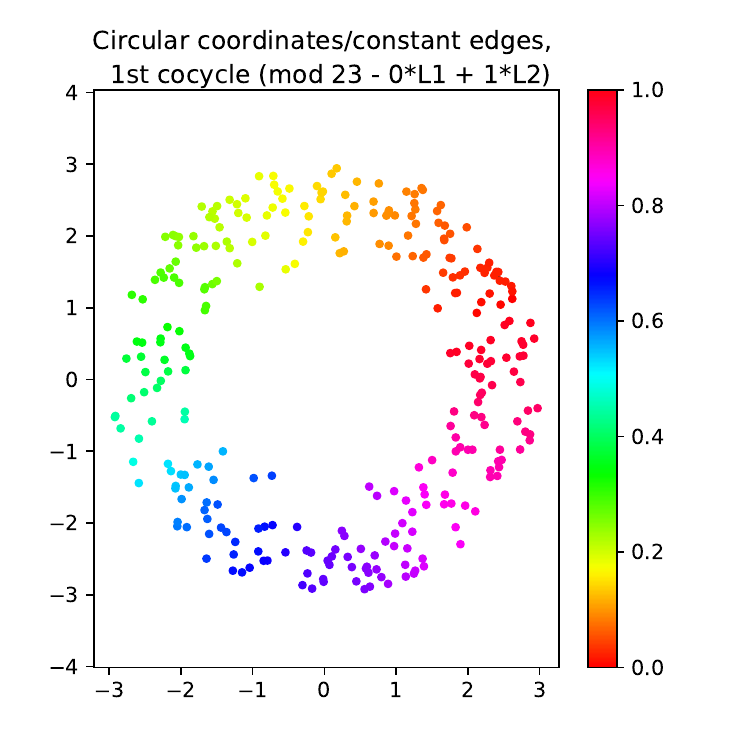}  &\tabularnewline
\includegraphics[height=5.7cm,page=10]{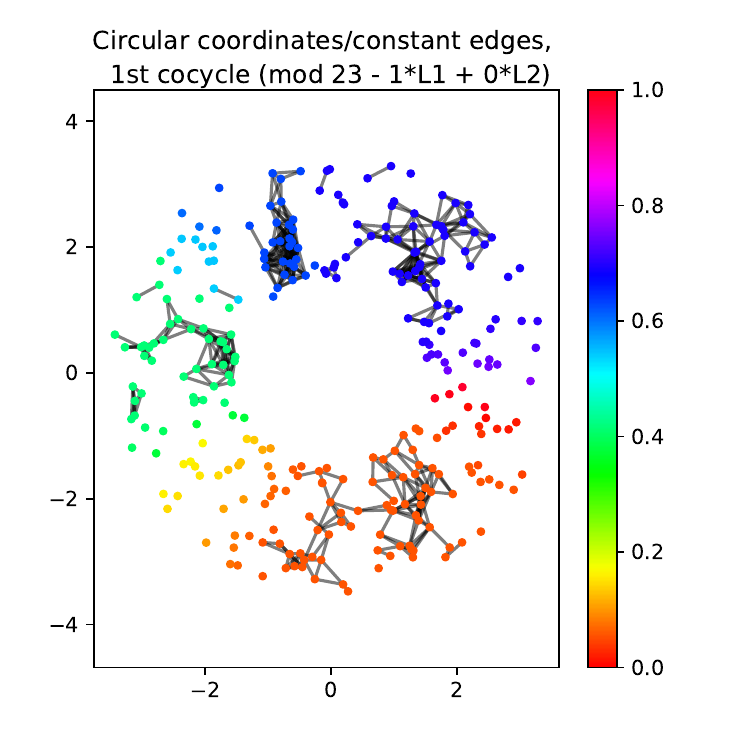}  & \includegraphics[height=5.7cm,page=24]{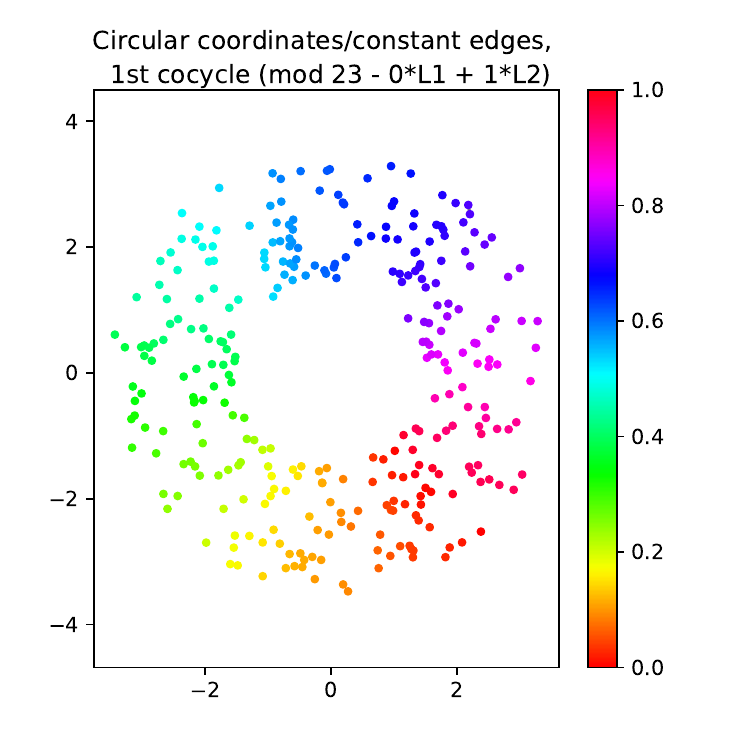}

& \tabularnewline
\includegraphics[height=5.7cm,page=10]{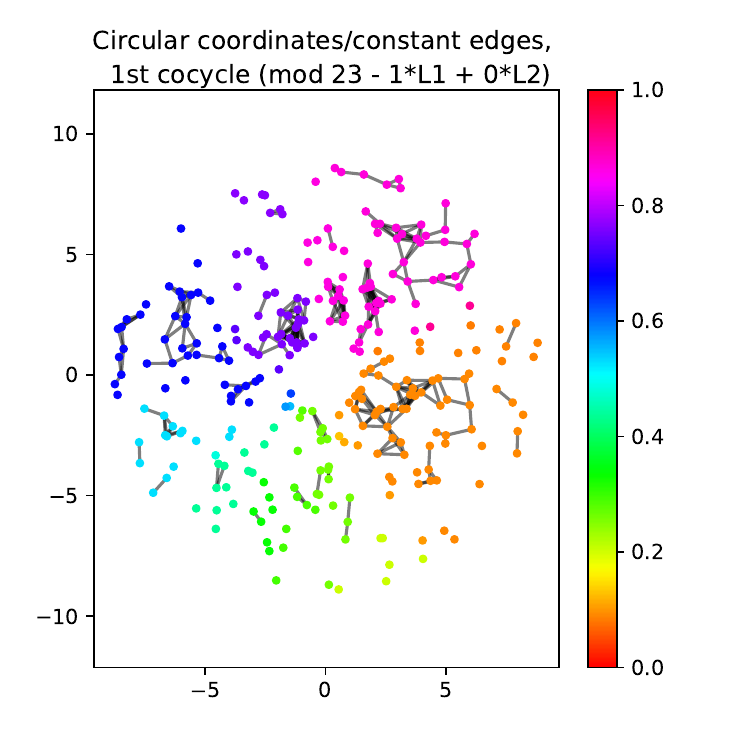}  & \includegraphics[height=5.7cm,page=24]{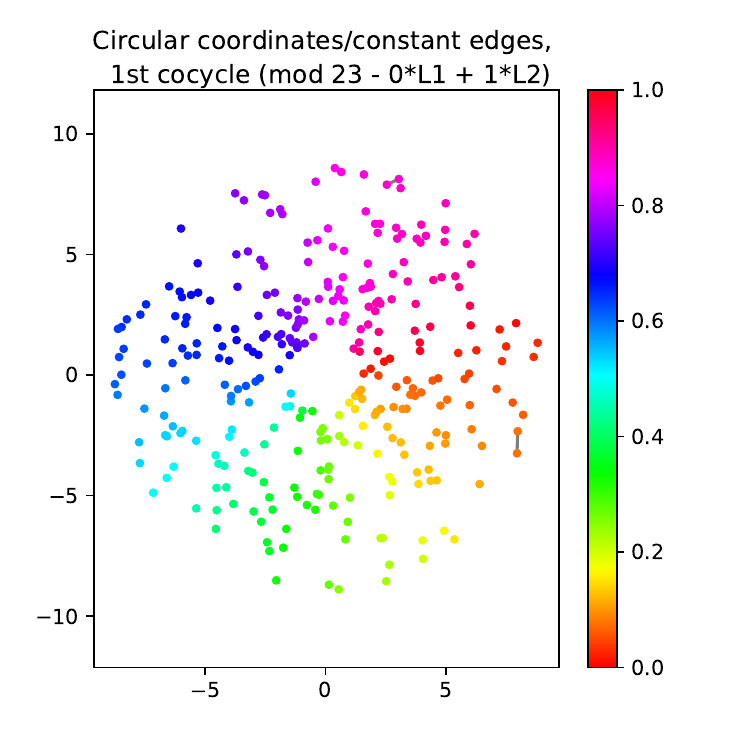}
\tabularnewline
\end{tabular}

\caption{\label{fig:Example-1:B}Example 1: The $L_{1}$ smoothed (first column)
and $L_{2}$ smoothed (second column) circular coordinates of the uniformly
sampled dataset from a ring with the same radius $R=1.5$ but different
widths $d=1,2,7.5$, corresponding to each row. The first and second columns correspond to $\lambda=0$ and
1, respectively.}
\end{figure}

\begin{figure}[t!]
\centering
\begin{adjustbox}{center}

\begin{tabular}{cccc}
$d=1$ & $d=2.5$ & $d=7.5$ &\tabularnewline
\midrule
\includegraphics[width=4cm]{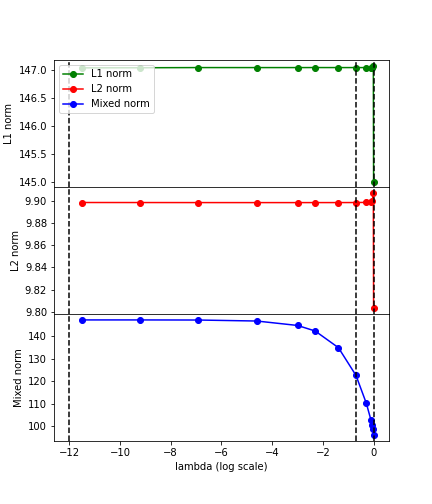} &
\includegraphics[width=4cm]{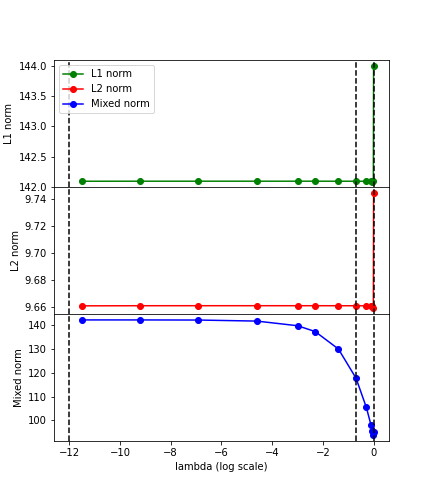} &
\includegraphics[width=4cm]{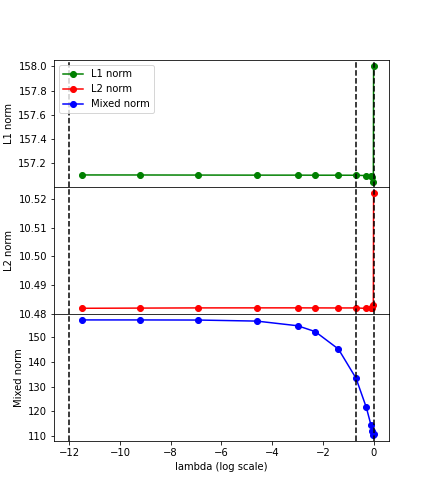} \tabularnewline
\end{tabular}
\end{adjustbox}
\caption{\label{fig:Example-sensitivity}Functional norms of varying $\lambda$ coefficient on Example 1: The $L_1$ (first row), $L_2$ (second row), and mixed norm (third row) for smoothed circular coordinates functions optimized with different choices of $\lambda$ as in (\ref{eq:cohomologous opt - L1}). The coordinates are computed for the uniformly
sampled dataset from a ring with the same radius $R=1.5$ but different
widths $d=1,2,7.5$, as in Figure \ref{fig:Example-1:B}, corresponding to each column. We also use black vertical dashed lines to delineate the $\lambda=0,0.5,1$ on the log scale.}
\end{figure}

In the next set of results shown in Figure \ref{fig:Example-1:B} we study the effect of the size of the hole defining the cocycle. We vary
the width $d$ of the ring, but keep the inner radius $R$ constant.
We can see that the inner hole of the ring becomes relatively smaller compared to
the rest of the ring area as $d$ increases. The difference between
the distribution of constant edges becomes less obvious as $d$ increases. This is expected and is due to the fact that as the width $d$ increases, the sampled points from the ring become more similar to a set of points uniformly distributed over a disc. The contrast between $L_{1}$ and $L_{2}$ penalties in the circular coordinates  vanishes as the width $d$ of the ring grows (more like a uniform distribution over the disc).
It is clear from Figure \ref{fig:Example-1:-Ring} that the $L_{1}$ penalty tends to produce more constant edges in the coordinate representation than the $L_{2}$ penalty.

\FloatBarrier

\subsection{\label{subsec:Example-2:-Double}Example 2: Double ring}

For double rings (a.k.a. double annulus) in $\mathbb{R}^2$, we consider two rings with width $d=1$ and inner radius $R=1.5$. These two rings are centered at $(-2,0)$
and $(2,0)$, respectively, with a nontrivial intersecting region. We sample 100 points in total,
50 points from each ring. We sample uniformly in the square $[0,1]\times[0,2\pi]$,
and then we use a standard parameterization as the previous example
onto two different rings areas in $\mathbb{R}^{2}$. Since the
topology of this example has Betti number 2, we choose two most significant 1-cocycles and obtain two circular coordinates
for this dataset.

\begin{figure}[t!]
\centering
\begin{adjustbox}{width=\textwidth}
\begin{tabular}{cccc}
cycle 1 & cycle 1 correlation plot & cycle 2 & cycle 2 correlation plot\tabularnewline
\midrule
\includegraphics[height=5cm,page=10]{figs/Example2_output_edit.pdf}  & \includegraphics[height=5cm]{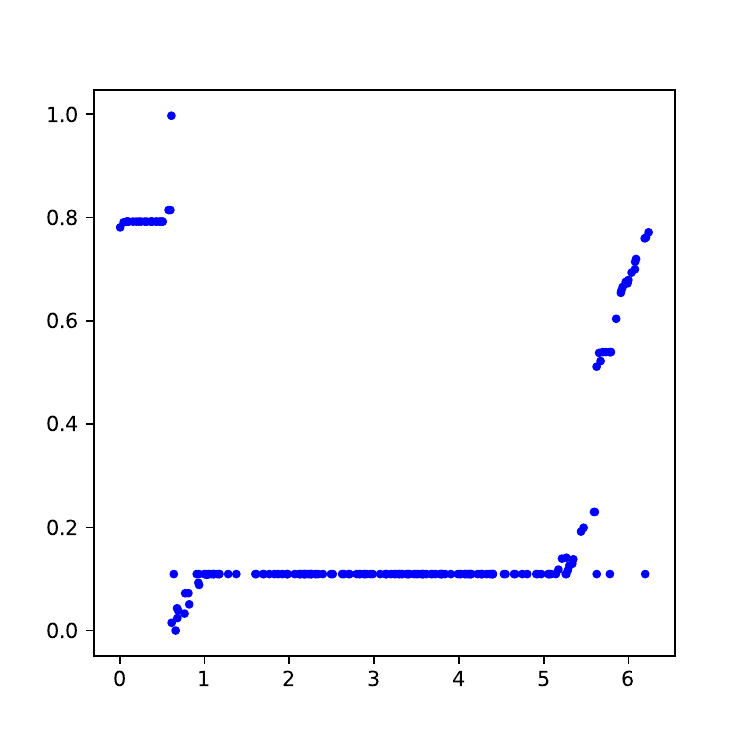} & \includegraphics[height=5cm,page=14]{figs/Example2_output_edit.pdf} & \includegraphics[height=5cm]{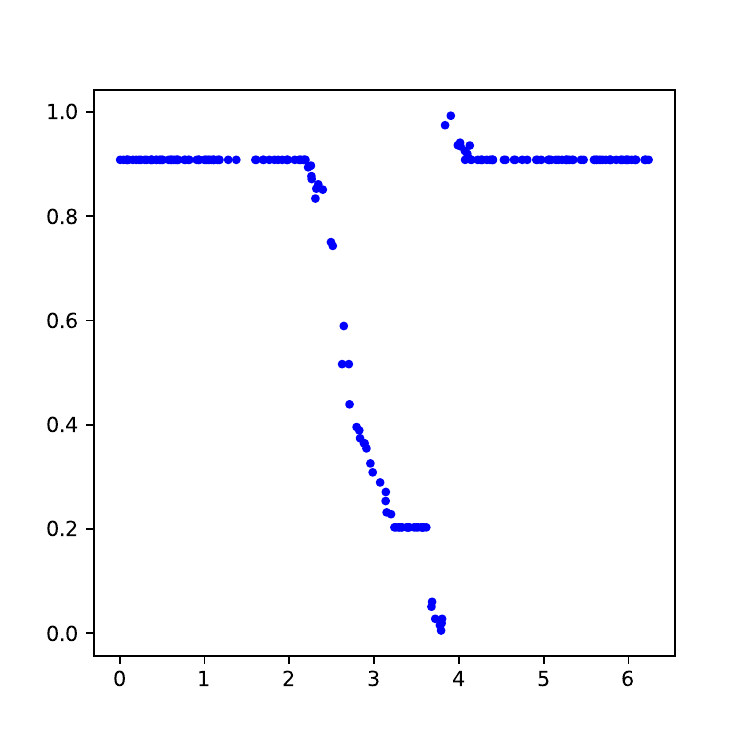}
\tabularnewline
\includegraphics[height=5cm,page=21]{figs/Example2_output_edit.pdf}  & \includegraphics[height=5cm]{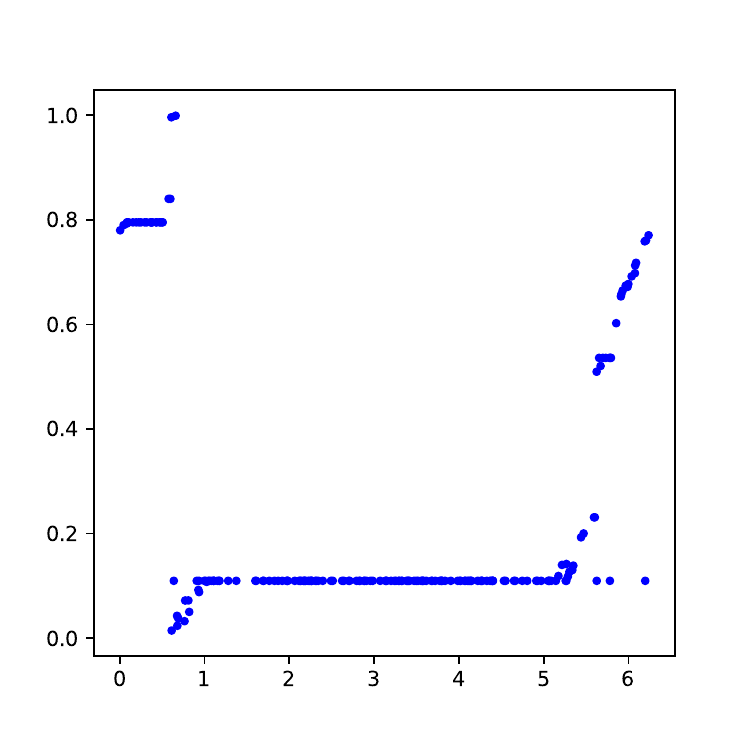} & \includegraphics[height=5cm,page=25]{figs/Example2_output_edit.pdf} & \includegraphics[height=5cm]{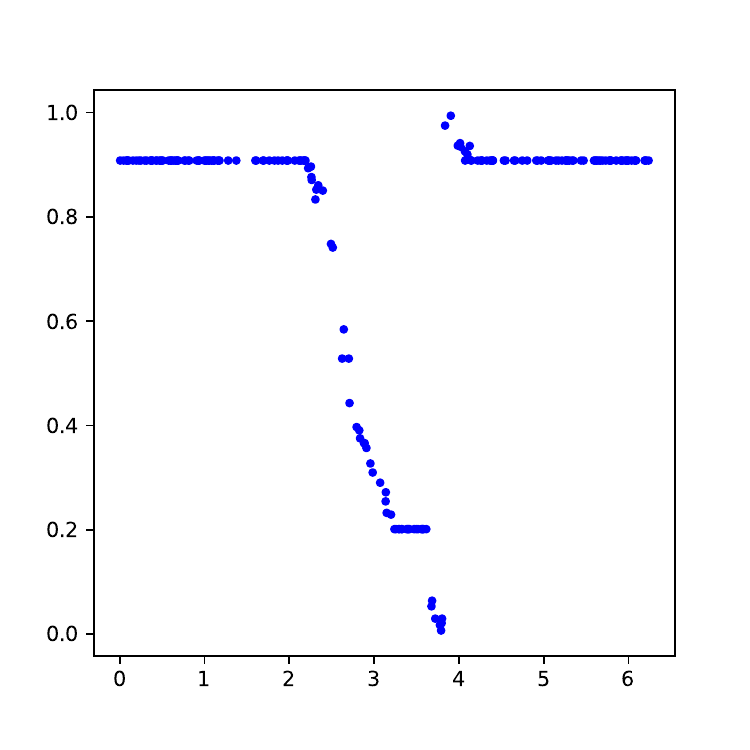}
\tabularnewline
\includegraphics[height=5cm,page=32]{figs/Example2_output_edit.pdf}  & \includegraphics[height=5cm]{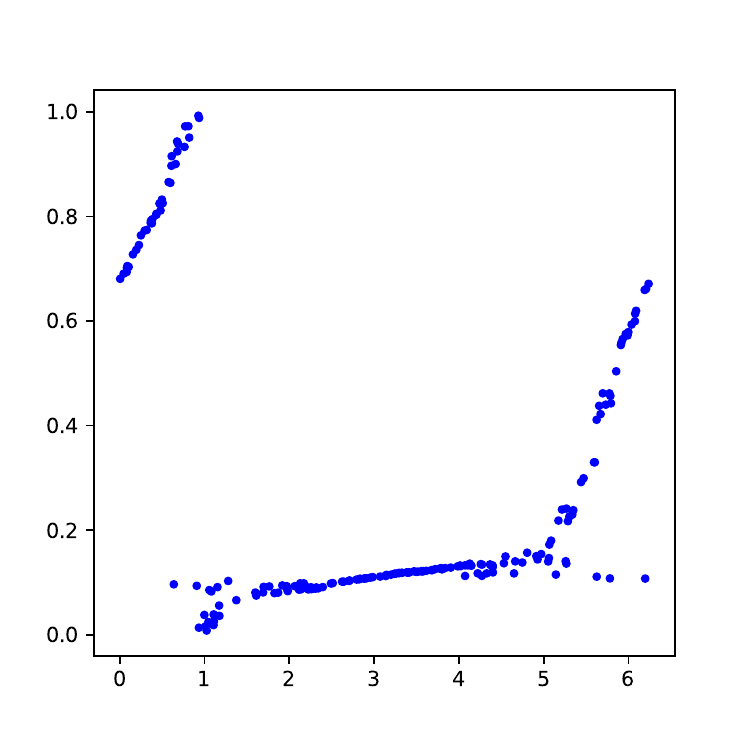} & \includegraphics[height=5cm,page=36]{figs/Example2_output_edit.pdf} & \includegraphics[height=5cm]{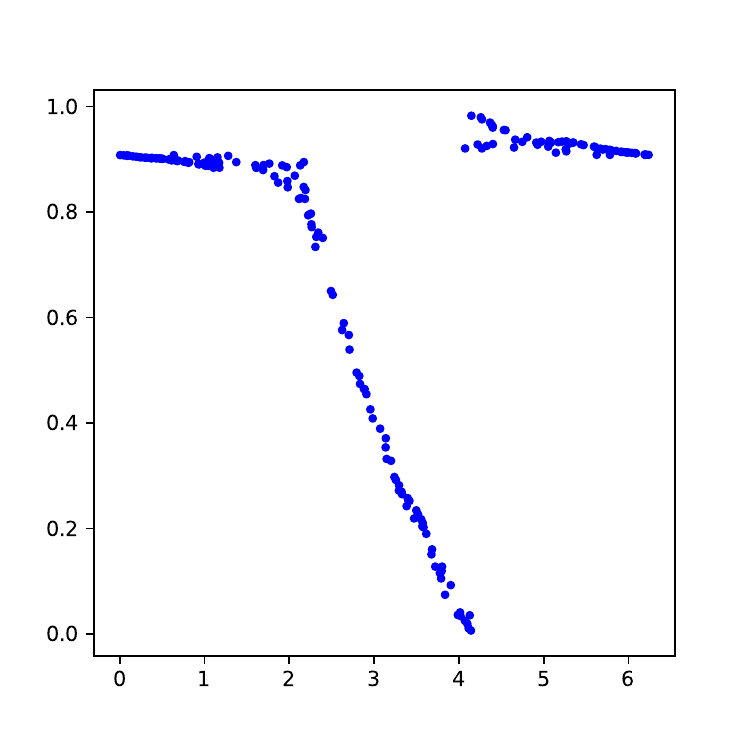}
\tabularnewline
\end{tabular}
\end{adjustbox}
\caption{\label{fig:Example-2}Example 2: The $L_{2}$ smoothed and generalized
penalized circular coordinate (displayed in different rows) of the
uniformly sampled dataset ($n=100$) from double rings, both with
inner radius $R=1.5$ and width $d=0.5$. The first, second, and the
third row correspond to $\lambda=0,0.5$, and 1, respectively.}
\end{figure}

Each cocycle leads
to one set of coordinate values for every point in
the sample. We can observe that each individual coordinate captures a different topological feature by showing non-constant
 coordinate values around the feature (i.e., on the boundaries of one of the rings).
 It is again observed that once we replace the $L_{2}$ penalty used in the circular coordinates, the number of constant edges increases. From our experience, this observation holds when there are multiple more complicated topological structures represented by 1-cocycles in the dataset.
Another observation is that once we
deviate $\lambda$ from $1$ in the generalized penalty function  $(1-\lambda)\|\cdot\|_{L_{1}}+\lambda\|\cdot\|_{L_{2}}$,
the ``outburst'' of the number of constant edges occurs quickly.
This can be observed by comparing different rows in Figure \ref{fig:Example-1:-Ring}
and \ref{fig:Example-2}. Therefore, there is little difference between the case $\lambda=0.5$ and $\lambda=0$. It is also one of the causes that lead to
the numerical instability of the optimization problem (\ref{eq:elastic net})
and (\ref{eq:cohomologous opt - L1}).

\FloatBarrier

\subsection{\label{subsec:Example-3:-Dupin}Example 3: Dupin cyclides}

Dupin cyclides are common examples of surfaces in $\mathbb{R}^3$ with nontrivial topological structures, since they have
different kinds of topologies as the parameter varies \cite{berberich2008exact}.
Here we focus on the case known as ``pinched torus''. For the pinched
torus of radii $r=2, R=1.5$, we sample 300 points $(x,y)$ at random
in the square $[0,2\pi]\times[0,2\pi]$  and parameterize them with
\[
(x,y)\mapsto\left(\left(r+\sin\frac{x}{2}\cdot\cos y\right)R\cdot\cos x,\left(r+\sin\frac{x}{2}\cdot\cos y\right)R\cdot\sin x,R\cdot\sin\frac{x}{2}\cdot\sin y\right),
\]
to map the points $(x,y)$ onto a surface in $\mathbb{R}^{3}$.


In \cite{robinson2019} it was conjectured that in the case of stratified manifolds -- such as the pinched torus -- the changes in a circular coordinate with a generalized penalty would concentrate near the strata.
That way, one could use the circular coordinates to locate these kinds of features.

\begin{figure}[t!]
\centering

\begin{tabular}{cc}
circular coordinates & correlation plot\tabularnewline
\hline
\includegraphics[height=5.7cm,page=10]{figs/Example_pinched__torus_output_edit.pdf}  & \includegraphics[height=5.7cm]{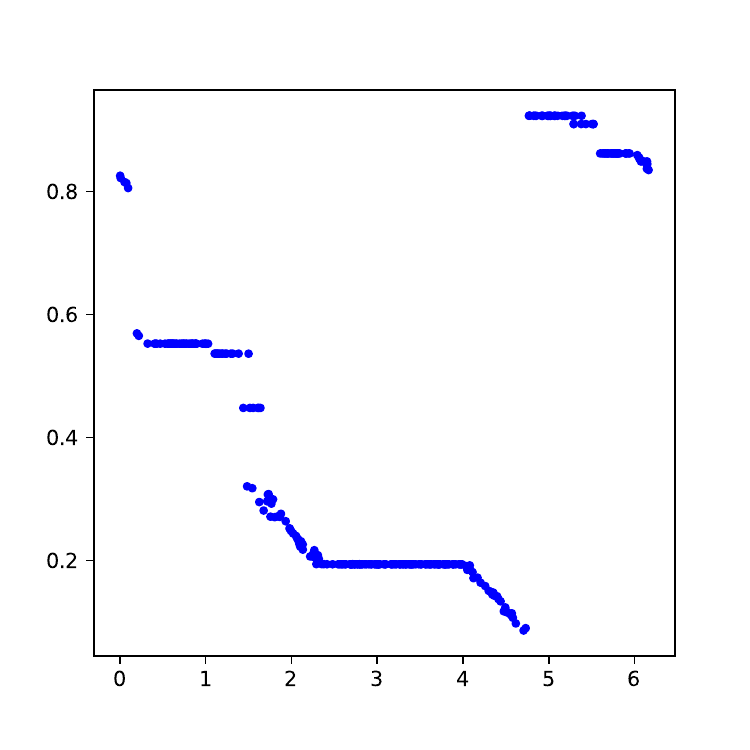}
\tabularnewline
\includegraphics[height=5.7cm,page=17]{figs/Example_pinched__torus_output_edit.pdf} & \includegraphics[height=5.7cm]{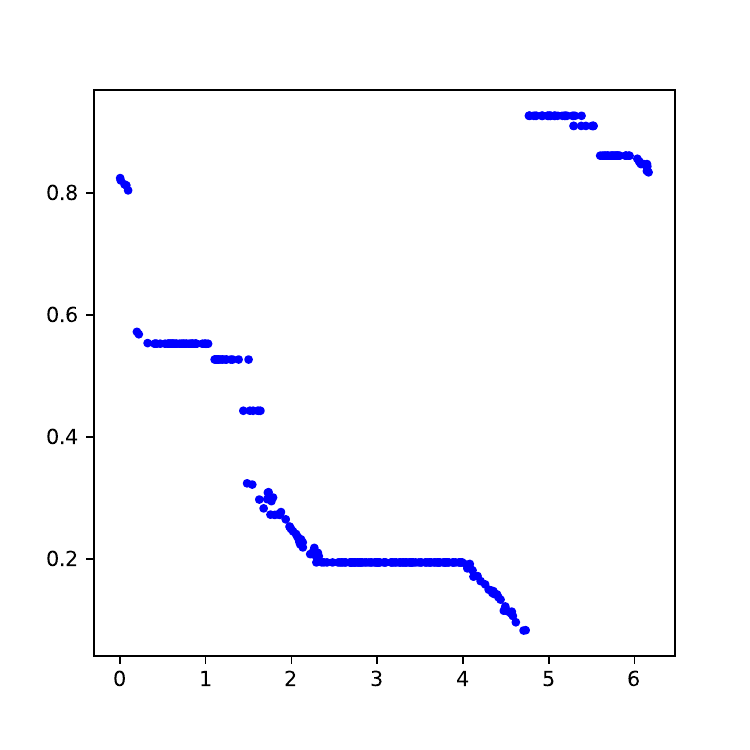}
\tabularnewline
\includegraphics[height=5.7cm,page=24]{figs/Example_pinched__torus_output_edit.pdf} &
\includegraphics[height=5.7cm]{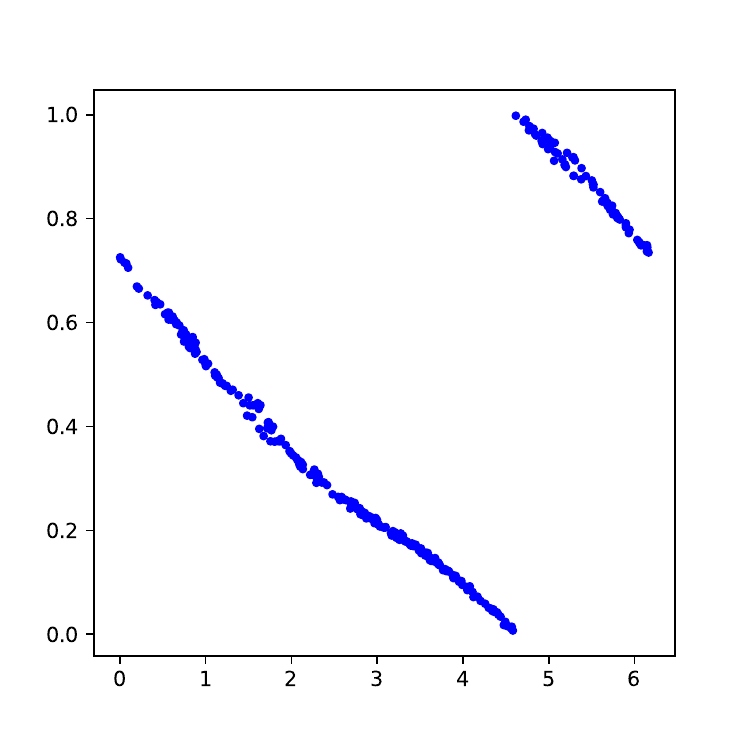}
\tabularnewline

\end{tabular}

\caption{\label{fig:Example-3}Example 3: The $L_{2}$ smoothed and generalized
penalized circular coordinate (displayed in different rows) of the
uniformly sampled dataset ($n=300$) from Dupin cyclides (a.k.a. pinched
torus). The first, second, and the third row correspond to $\lambda=0,0.5$,
and 1, respectively.}
\end{figure}

To evaluate the conjecture, we have looked at two different sampling schemes for the pinched torus: in Figure~\ref{fig:Example-3}, we sample the parameter space $[0,2\pi]\times[0,2\pi]$ uniformly at random, and in Figure~\ref{fig:Example 6} we use the volume uniform sampling scheme that we will describe in Section~\ref{subsec:Volume-uniform-sampling-scheme}.

In Figure~\ref{fig:Example-3}, we see that instead of the \emph{change} concentrating near the pinch point, we have a large constant region covering the pinch point.
On reflection, this is a reasonable outcome: by sampling the parameter space uniformly, we get a higher density of sample points near the pinch point.
With the higher density comes a larger amount of (short) edges, and thus the optimizer is guided away from this region towards more sparsely populated regions to capture the same amount of change with the smallest possible number of edges.

When using volume uniform sampling however -- adjusting the sampling scheme to compensate for the area distortion between the parameter space and the surface -- a different picture emerges.
In Figure~\ref{fig:Example 6} we see that regardless of whether we use $L_2$ penalties or a sparsified regime, the change concentrates near the pinch point and any constant edges -- if they occur -- are placed far from the pinch point.

The conclusion we are led to by these observations is that 
Robinson's conjecture seems contradicted by parameter uniform sampling, but in the examples computed in Appendix~\ref{subsec:Volume-uniform-sampling-scheme} the opposite effect happens: constant edges are placed far from the pinch point and the change concentrates near the pinch point.

\FloatBarrier

\subsection{\label{subsec:Uniform summary}Effect of generalized penalty functions in circular coordinates}
As we have seen in the simulation examples above, the generalized penalty function we proposed in (\ref{eq:cohomologous opt - L1}) produces more constant edges in a small region of the space than the classical $L_2$ based penalty. Through the example application on the ring (Section \ref{subsec:Example-1:-Ring}) we show the different behavior of $L_2$ and the generalized penalty and how the distribution of constant edges evolves as the inner radius increases (i.e. from a sparse narrow ring to a disc). In the double ring example (Section \ref{subsec:Example-2:-Double}) we repeated a similar experiment considering multiple 1-cocycles, and reveal how circular coordinates from different 1-cocycles can indicate different features in the manifold. The last example, using the Dupin cyclide as the underlying manifold (Section \ref{subsec:Example-3:-Dupin}), allowed us to  address a conjecture by \cite{robinson2019} speculating changes in circular coordinates for stratified manifolds. Indeed, we found a concentration of constant edges around the pinched point.

In short, we have shown how the generalized penalty function will help us identify topological features. The circular coordinates with generalized penalty function would \label{properties1}
\begin{enumerate}
\item Enforce more roughness in terms of jumps in circular coordinate function values as a solution to (\ref{eq:cohomologous opt - L1}), thus generating more constant edges on the region of the dataset with no topological variation.
\item For dataset without jumps, the circular coordinates with generalized penalty function produce non-concentrating constant edges, similar to the $L_2$ penalty (Fig. \ref{fig:Example-1:B}).
\end{enumerate}

In this section, we avoid discussing the distributions of constant edges in our simulations above, since these seem to be related to the sampling scheme used to produce the samples. In Appendix \ref{subsec:Volume-uniform-sampling-scheme}, we repeat the simulations with a volume uniform sampling scheme to look further into the effect of the sampling scheme on the resulting concentration of constant edges.

\section{\label{sec:Real-data-analysis}Real data analysis}

In the previous simulation studies, we focused on locating the nontrivial topological features in the underlying manifold $M$ using varying circular coordinate values. Our observations show that by
using a generalized penalty function, we can obtain a coordinate representation while preserving the topological features of the high-dimensional dataset. (See page \pageref{properties1} and \pageref{properties2})

We now want to show the performance of the circular coordinate representation with a generalized penalty function, using real datasets.
Unlike simulation datasets, where we know the true underlying topology of $M$ and can easily isolate the relevant 1-cocycles from topological noise, in real datasets, where we do not know if significant 1-cocycles exist, making this distinction can be tricky. We need to try a different number of 1-cocycles to decide how many of them are significant and we want to retain for computing circular coordinates.
Another important difference is that both real datasets we consider here are high-dimensional in nature, while the simulation datasets are in $\mathbb{R}^2$ or $\mathbb{R}^3$. 

\subsection{Sonar record}

This raw sonar record dataset was collected by \cite{robinson2012multipath} using a sonar device to record ceiling fan frequencies. It is argued in the paper to be especially suitable for circular coordinate analysis since the multipath components of the data show quasi-periodicity. This real data analysis is dedicated to showing the contrast of the analysis result between the classical $L_2$ penalty against the generalized penalty on a real-world dataset.

The data
is a $175\times1300$ rectangular matrix whose columns represent sonar
pulses and rows represent particular range-bins (i.e., distance to
the sonar). The sonar records are collected from three different setups
of the ceiling fan: rotating counterclockwise, with frequency of 1/3
Hz; rotating counterclockwise, with frequency 1 Hz and rotating clockwise,
with frequency 1 Hz (i.e., coded as collection 3, 4, and 5, respectively.).

The dataset comes as a natural high-dimensional dataset due to its data collection procedure, and it has some level of periodicity in the data generating mechanism. The periodicity in this dataset also produces the cocycles we need to apply the circular coordinate methodology. Since the fan speed is not exactly
constant during the data collection procedure, \cite{robinson2012multipath} attempts to identify the parameterization of the fan's rotation from the data using circular coordinates. Following a suggestion from \cite{robinson2019}
to drop the near-in clutter (i.e., places near the sonar)
noises, we drop the first 250 columns
of the matrix and use circular coordinates (with $L_{2}$ and generalized penalty functions) to investigate the quasi-periodicity exhibited across different distance-bins.

\begin{figure}[t]
\centering

\includegraphics[height=12.7cm]{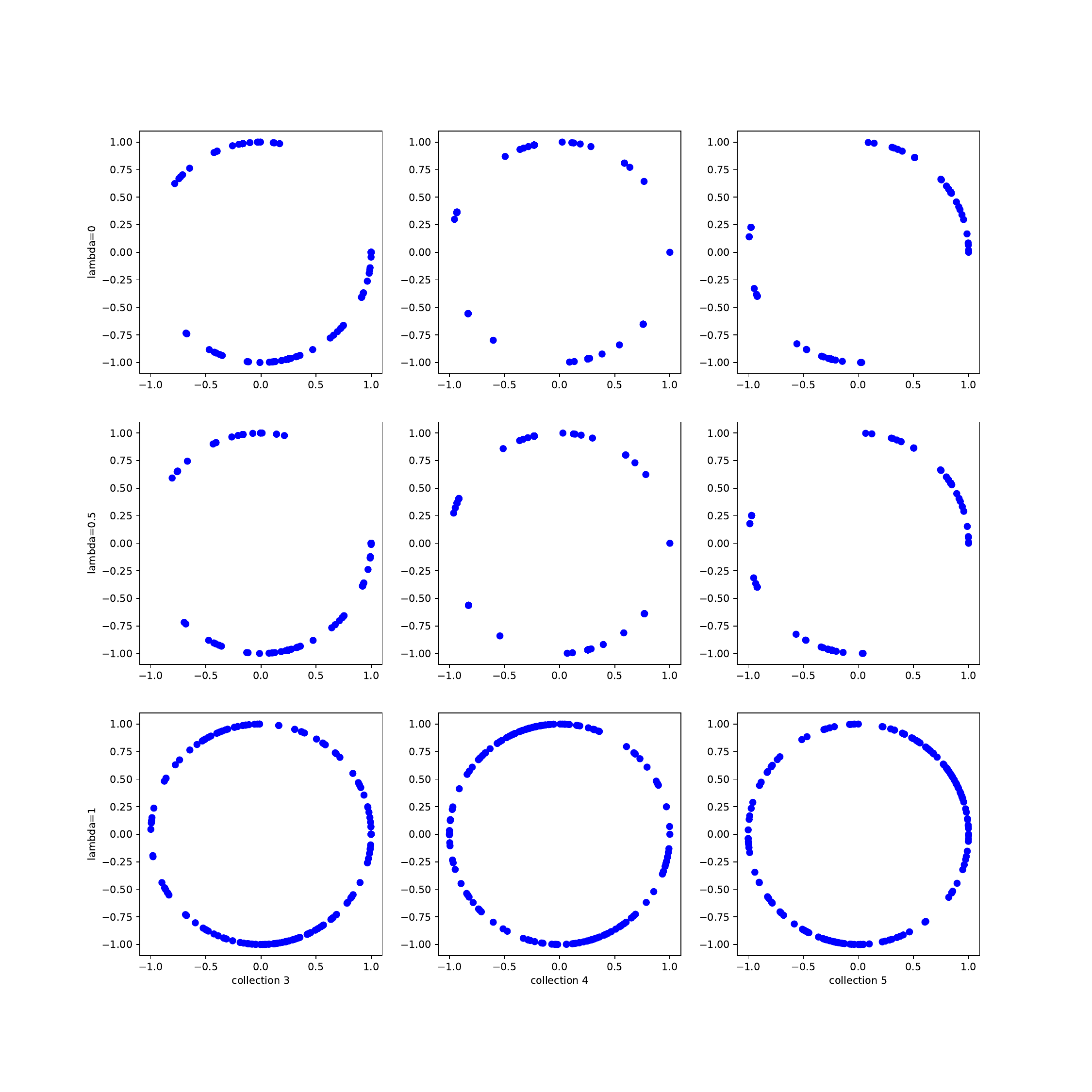}

\caption{\label{fig:Example fan 1} The $S^{1}$ representation obtained from
the circular coordinate representation under different penalty functions.
The first, second, and the third row correspond to $\lambda=0$, $0.5$,
and $1$, respectively.}
\end{figure}

\begin{figure}
\begin{adjustbox}{center}

\begin{tabular}{ccc}
collection 3 & collection 4 & collection 5\tabularnewline
\hline
\includegraphics[width=4cm,page=1]{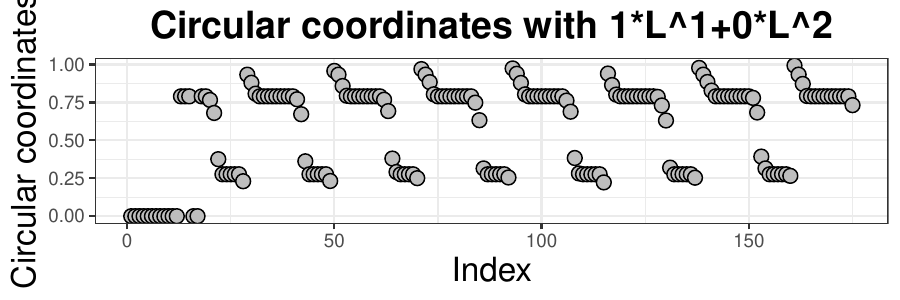} & \includegraphics[width=4cm,page=1]{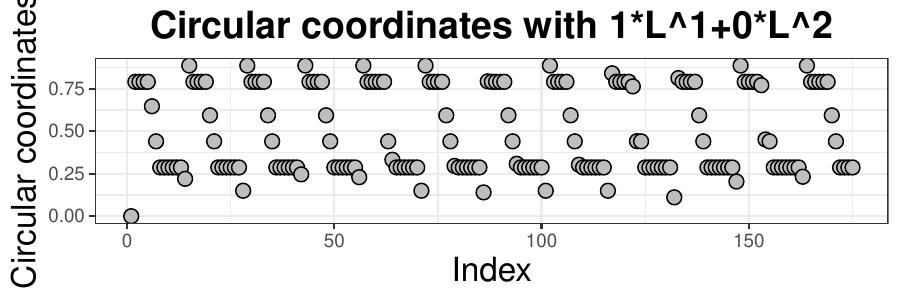} & \includegraphics[width=4cm,page=1]{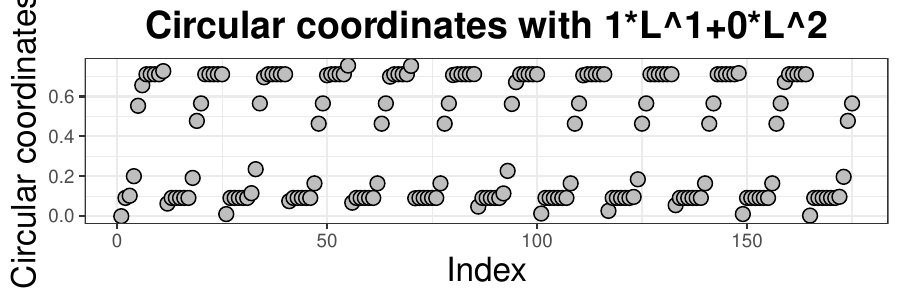}\tabularnewline
\includegraphics[width=4cm,page=2]{figs/collection_3_plot} & \includegraphics[width=4cm,page=2]{figs/collection_4_plot} & \includegraphics[width=4cm,page=2]{figs/collection_5_plot}\tabularnewline
\includegraphics[width=4cm,page=3]{figs/collection_3_plot} & \includegraphics[width=4cm,page=3]{figs/collection_4_plot} & \includegraphics[width=4cm,page=3]{figs/collection_5_plot}\tabularnewline
\end{tabular}
\end{adjustbox}
\caption{\label{fig:Example fan 2} The $L_{2}$ smoothed and generalized penalized
circular coordinates (displayed in different rows) of the three collections
of fan frequency dataset ($n=175$) from \cite{robinson2012multipath}
plotted against indices (equivalent to the distances of distance-bins). The first, second, and the third row correspond
to $\lambda=0$, $0.5$, and 1, respectively. The circular coordinates with generalized penalty function are much sparser compared to the coordinates associated with the $L_2$ penalty function, which means that our method captures the periodic pattern better.}
\end{figure}

We computed the circular coordinates using Vietoris-Rips
complexes in $\mathbb{R}^{1050}$ and choose the most significant 1-cocycle only. In Figure \ref{fig:Example fan 2},
we select significant cocycles with persistence greater than the significant threshold
$\eta=1$.
The result shown in Figure \ref{fig:Example fan 1} displays circular coordinates after the coordinates are embedded into $S^{1}$.
In this visualization, the quasi-periodicity of the frequencies against distances is represented as the (quasi-)periodicity in circular coordinates. In addition, we can observe from the plot of circular coordinates against indices
of the data (equivalent to the distances of distance-bins) in Figure~\ref{fig:Example fan 2} that the generalized penalty function will
give us more constant coordinate values. We can see that the jumps are highlighted under the $L_{1}$ penalty and the change of signals is more abrupt. Furthermore, this abrupt pattern reveals quasi-periodicity qualitatively better: in Figure~\ref{fig:Example fan 2}, the patterns in circular coordinates among different collections are more distinguishable under the $L_{1}$ penalty than under the $L_{2}$ penalty (different columns). Within the same collection, $L_1$ would show the transition between periods better while $L_2$ would display the in-period variations better (different rows). Using the abruptly varying pattern, the qualitative differences between different collections are better represented.

To sum up, this sonar record example shows how circular coordinates
with the $L_{2}$ and generalized penalty functions differ in the
representation of sonar signals. The $L_1$ and generalized penalty function identifies the periodic pattern qualitatively better in the final signal representation compared to the $L_2$ penalty function. 


\FloatBarrier

\subsection{\label{Voting Data}Congress voting}

The dataset we are analyzing is a U.S. congress voting record (1990-2016)
dataset collected by \cite{vejdemo2012topology}. The congress
voting dataset has been studied and it is observed to have nontrivial
connectivity features along with many extreme circles existing in the dataset. 
Therefore, the classical circular coordinates methodology has been applied to this voting record dataset to detect the evolution of voting patterns. For this real-world dataset, we focus on the difference between these two sets of coordinates instead of one coordinate alone. This real data analysis is dedicated to using the comparison information from different penalties.

Datasets from different opening years come in the form of an $m\times d$ matrix, consisting of $m$ congressmen/women
voting on $d$ issue bills. Each row of this data matrix stores a voting
record of a specific congressman/woman during a certain opening year,
and each column of this data matrix denotes a bill on a certain issue.
When a congressman/woman votes ``Yes'' for a certain bill, we denote
it by 1; when a congressman/woman votes ``No'' for a certain bill,
we denote it by -1. We fill in 0 for all other outcomes other than ``Yes'' or ``No'' (including abstaining).

\begin{figure}[t!]
\centering
\begin{centering}
\begin{adjustbox}{width=\textwidth}
\begin{tabular}{cc}
$\lambda=0$ & $\lambda=1$\tabularnewline
\hline
\includegraphics[width=9cm,page=2]{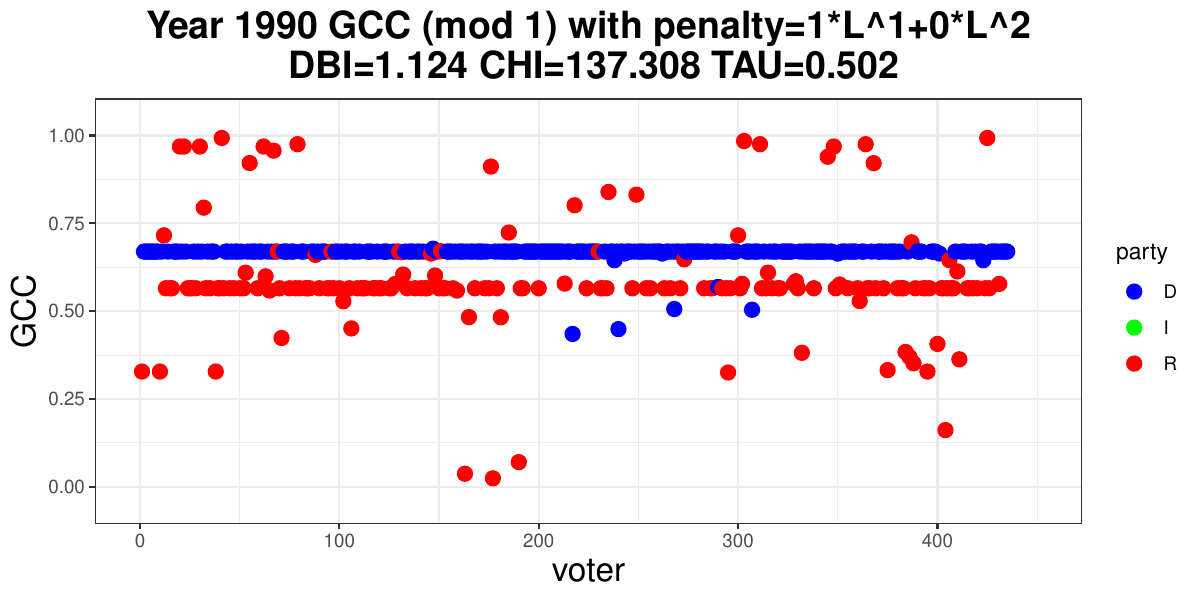} & \includegraphics[width=9cm,page=8]{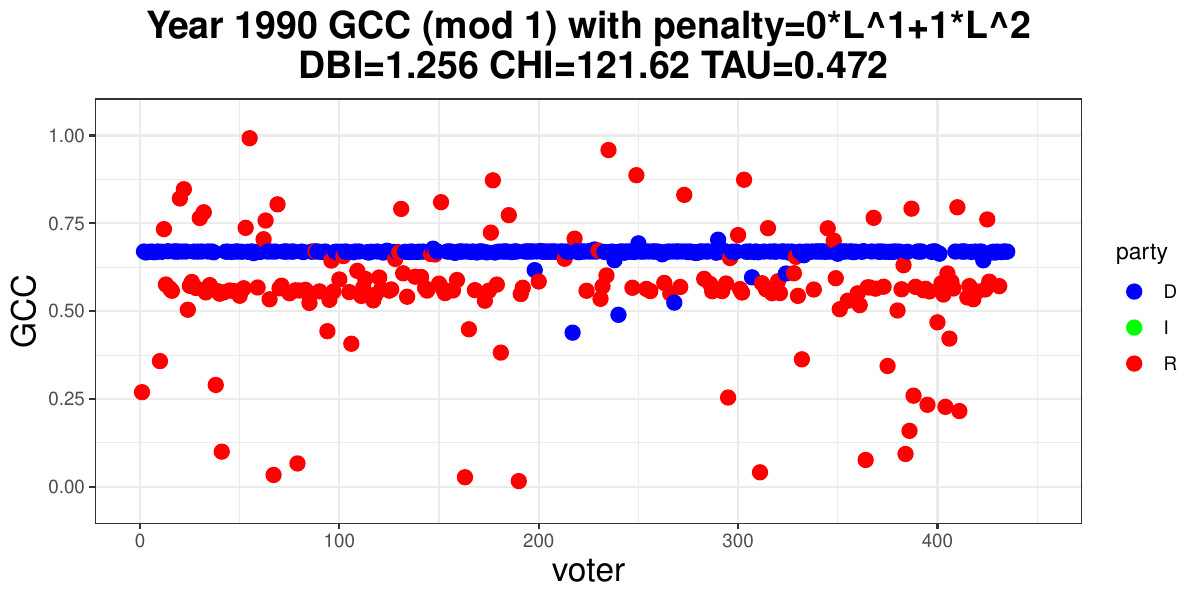}\tabularnewline
\hline
\includegraphics[width=9cm,page=11]{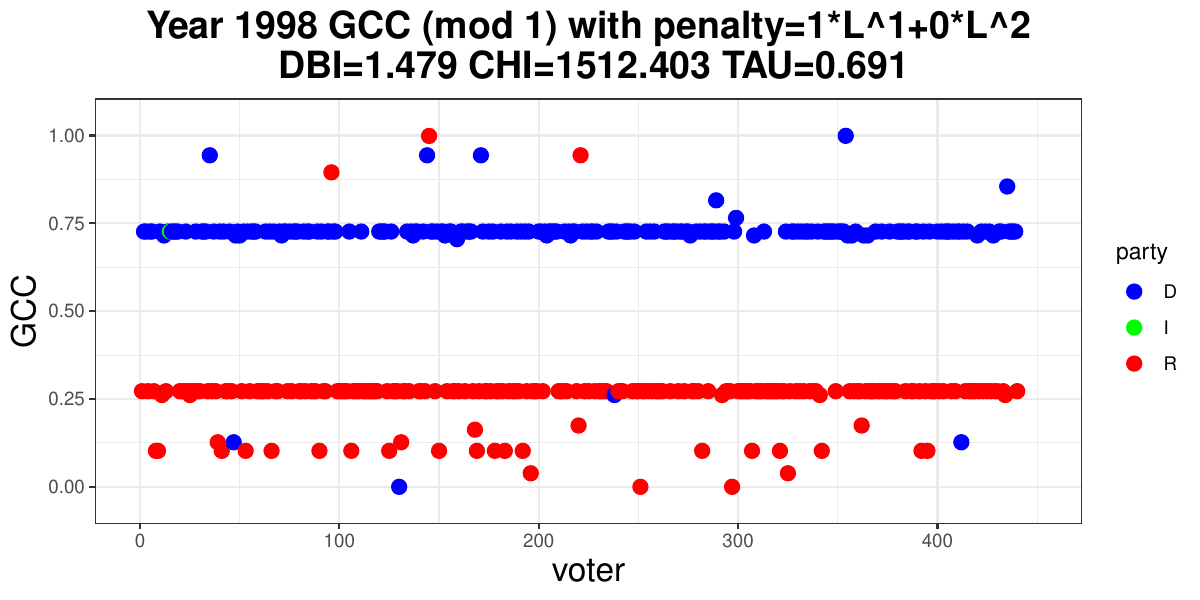} & \includegraphics[width=9cm,page=17]{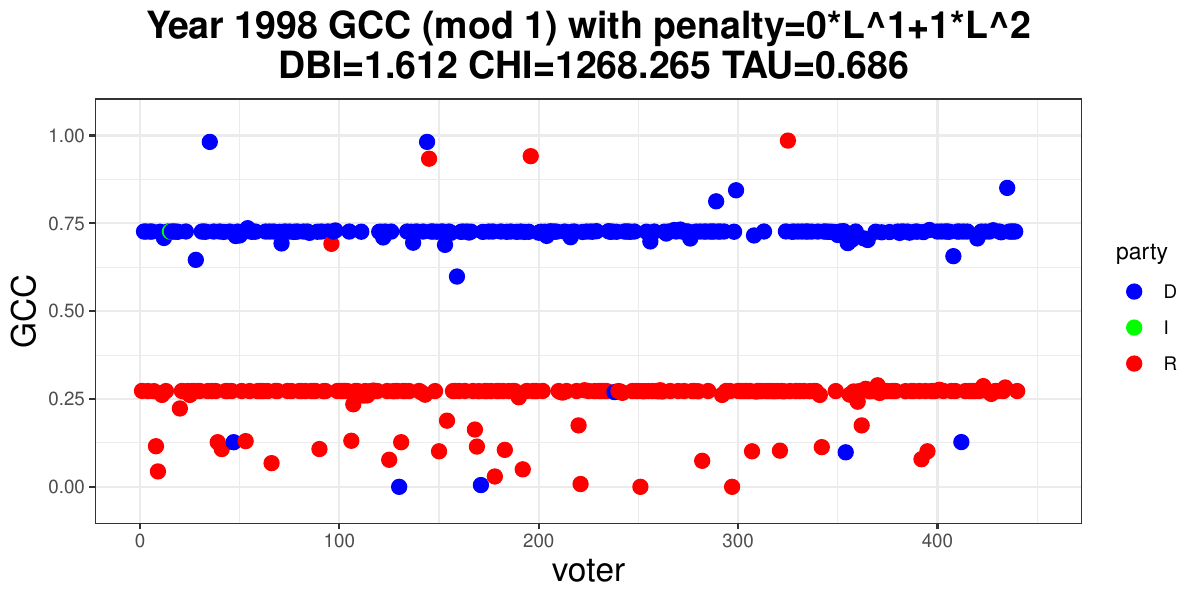}\tabularnewline
\hline
\includegraphics[width=9cm,page=20]{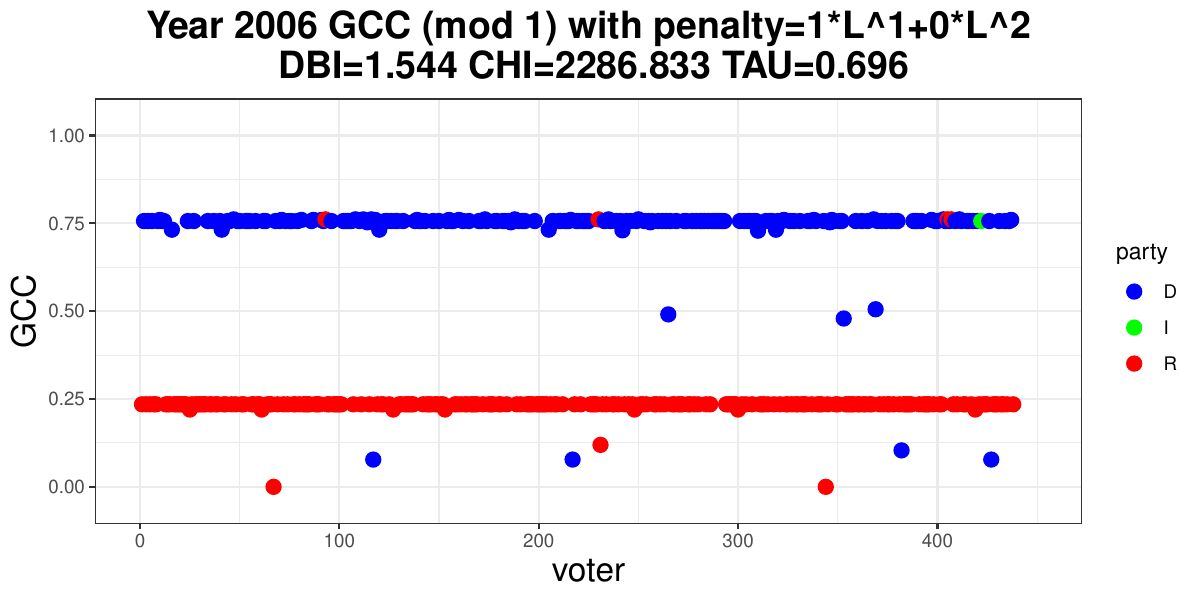} & \includegraphics[width=9cm,page=26]{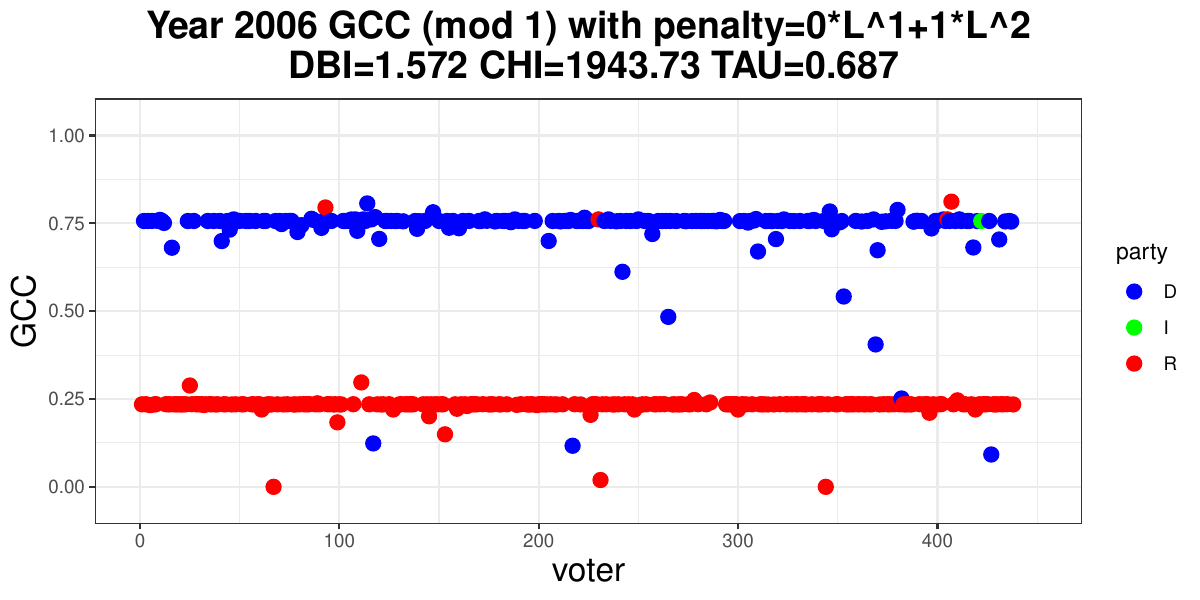}\tabularnewline
\end{tabular}
\end{adjustbox}
\par\end{centering}
\caption{\label{fig:Example Real Data} The $L_{2}$ smoothed and generalized
penalized (mod 1) combined circular coordinates among congressman/woman across party-lines. Each point represents a congressman/woman and the color represents party-lines. The circular coordinates are computed from congress voting datasets from years 1990, 1998, and 2006 (displayed in different rows). The
first and the second column correspond to $\lambda=0$ and $1$, respectively. We compute the cluster scores by mapping the combined circular coordinates (summed up by all 1-cocycles with persistence greater than 1) to $\mathbb{R}^2$ with the mapping $x\mapsto(\cos(2\pi x),\sin(2\pi x))$ to accommodate the circularity.
\newline }
\end{figure}

As described above, we computed the circular coordinates using Vietoris-Rips
complexes in $\mathbb{R}^{d}$. In Figure \ref{fig:Example Real Data},
we select significant cocycles with persistence greater than threshold
$\eta=1$, and compute the circular coordinates under
$L_{1}$ and $L_{2}$ penalty, respectively\footnote{We omit the elastic norm here, since it is almost identical to $L_{1}$ penalty on
this specific dataset.}.

In the subsequent analysis, we use combined circular coordinates under different penalties. That is, we simply sum up the coordinate values computed from each significant 1-cocycle and plot the combined coordinate values against the voter serial number in Figure \ref{fig:Example Real Data}. To quantify the performance of clustering based on circular coordinate values, we consider the Davis-Bouldin index (DBI, \cite{davies1979cluster}), Calinski-Harabasz Index (CHI, \cite{calinski1974dendrite}) and the classical tau score (TAU, \cite{vendramin2010relative}). Lower values of DBI and higher values of CHI and TAU imply better separation and partition between clusters, i.e., party-lines in the dataset.

In the 1990 voting data, circular coordinates with $L_{2}$ penalty cannot separate
party-line and produces a lot of noise and outliers while circular coordinates
with $L_{1}$ penalty separate parties relatively clearly. We can also observe
this from the consistently better cluster scores in the $L_1$ coordinates clustering results.

In the 1998 voting data, both
$L_{1}$ and $L_{2}$ penalties produce reasonably clear coordinate
separations. However, we observed that the $L_2$ penalized coordinate has a slightly larger within-party variation, which causes the DBI inflation. The TAU and CHI of $L_2$  are also worse than those of $L_1$ coordinates, although their performances are close for the data in this year.

In the 2006 voting data, the coordinate values
under both $L_{1}$ and $L_{2}$ penalties manage to maintain a clear visual difference between the two parties.
In contrast, $L_{2}$ penalty produces a worse CHI in this year 2006, which means that the within-cluster dispersion and the between-cluster dispersion under $L_{2}$. The difference in graphical representation is not obvious, but we can read from quantitative indices that $L_1$ produces slightly better clustering here in terms of DBI and TAU scores.

With the same voting dataset, we examined the quantitative performance measure given by the coranking
matrix proposed by \cite{lee2009quality}.
Circular coordinates
reduce the dataset from $d$ dimensions into a lower dimensional dataset
of dimension $k$, where $k$ is the number of chosen significant
cocycles. To compare circular coordinates as a dimension
reduction method against other NLDR methods, we specify the dimension
reduced data to live in $\mathbb{R}^{k}$. By doing this, all
dimension reduction methods perform a dimension reduction from $\mathbb{R}^{d}$
to $\mathbb{R}^{k}$. We can see that the coranking matrix of
circular coordinates has a very sharp block structure (See Appendix
\ref{sec:Comparison-with-other}), which is similar to t-SNE \cite{maaten2008visualizing}
and UMAP \cite{mcinnes2018umap}. Similar observations for dimension
reduced dataset by PCA and Laplacian eigenmaps \cite{belkin2003laplacian}
indicate that these two methods do not preserve the group separation
(or clustering) in the dataset across different parties
well.

To sum up, this congress voting example shows how circular coordinates
with the generalized penalty functions provide a better group separation
with topological information. 
We use the different behaviors of circular coordinates smoothed using $L_1, L_2$ norms to show the different levels of sparsity in the dataset. When observing real data, the contrast of $L_1, L_2$ norms reveals additional information. Specifically, for the voting data, since the sparsity exists in the voting records, it seems that $L_1$ norm provides consistently better clustering.
We know from our simulations in Section \ref{sec:Simulation-studies-under} that the generalized penalty leads to sharper changes of the coordinate values near topological features compared to $L_{2}$. This explains why, when the data is sparse, we find that the circular coordinates with generalized penalty separate the clusters better when evaluated by the cluster scores chosen above. In fact, $L_{1}$ is better under almost all scores. The voting data is really sparse in the sense that there are roughly $3^{300}$ possible voting outcome combinations, while each year only contains approximately 300 representatives' votes.

\FloatBarrier

\section{\label{sec:Discussion}Discussion}

\subsection{Conclusion}

Our contribution in this paper can be summarized in two parts. We first propose a novel topological dimension reduction method that 
can effectively represent  high-dimensional datasets with nontrivial cohomological structures.
We then explore the behavior of generalized penalty functions on simulated and real datasets and show how they can be applied in a nonstandard setting.

The circular coordinate \cite{de2011persistent} is a nonlinear dimension reduction method, which is capable of providing a topology-preserved low-dimensional representation of high-dimensional datasets using significant 1-cocycles selected from the persistent cohomology based on the dataset.

The circular coordinate representation depends on the penalty function used in the cohomologous optimization problem in the form of:
\begin{equation}
\bar{f}=\arg\min_f\{(1-\lambda)\|\bar{\alpha}\|_{L_{p}}+\lambda\|\bar{\alpha}\|_{L_{q}}\mid f\in C^{0}(\Sigma,\mathbb{R}),\bar{\alpha}=\alpha+\delta_{0}f\}.\label{eq:pq-net}
\end{equation}

Using $L_{1}$ penalty function results in a more locally constant circular coordinate representation as seen in Section \ref{sec:Cohomologous-optimization-proble}.

Circular coordinate representation is also an effective visualization
tool for high-dimensional  datasets as we have seen in Section \ref{sec:Simulation-studies-under} and \ref{sec:Real-data-analysis}. Circular coordinates come with
a natural visualization as points on a circle or a torus, and we can
use the varying coordinate values to locate important
geometric features. The set of coordinates retains the topological information
of a dataset, and it can handle the sparsity in high-dimensional data with the concentration of constant edges near a small region, achieved via generalized penalty functions.

As we have seen in Section \ref{sec:Real-data-analysis}, the analysis of the sonar record and congress voting
supports our intuition that the circular coordinates reflect important features of the high-dimensional dataset relatively well. As a dimension reduction
technique, it is also effective in discovering (quasi-)periodicity (sonar records) or preserving clustering
structures (congress voting) in high-dimensional datasets, compared to other existing nonlinear dimension reduction methods (See Appendix \ref{sec:Comparison-with-other}).


In conclusion, we provide a novel method of nonlinear dimension
reduction and visualization method, namely, circular coordinate representation
with a generalized penalty. Our method comes with explicit roughness control in terms of generalized penalty functions and extends the
circular coordinates representation \cite{de2011persistent}.
We propose to use general penalty functions
to penalize the circular coordinates.
This extended procedure
preserves and facilitates the detection of topological features
using constant edges in a high-dimensional dataset. Although
the harmonic smoothness of coordinates under $L_{2}$ is not preserved,
the locally constant and abrupt pattern of circular coordinates by generalized penalty provides rich information of the topological structure of the dataset, such as (quasi-)periodicity or clustering structure.
Our generalization is motivated
by statistical considerations and it reveals how the idea of generalized penalty functions in statistics can be applied to accommodate for analyzing datasets. 

\subsection{Future works}
On the one hand, it would be interesting to explore other kinds of penalty functions already established in a regression setting.
On the other hand, it would also be important to explore whether the circular coordinates with generalized penalty can be helpful in model selection.
In dimension reduction, sparsity can be studied further with a view of robust statistics using a geometric induced loss function \cite{ronchetti2020main, luo_li_SPCA_2020}.
This line of research is motivated by the statistical literature on generalized penalty functions \cite{hastie2015statistical}.

Beyond the $S^1$ coordinate functions, it is of interest to explore whether the idea of penalized smoothing could be extended to coordinate functions with values in a general topological space other than $S^1$. In this direction, we want to explore the idea of generalized penalty functions with Eilenberg-MacLane coordinates, of which $S^1$ coordinate is a special case \cite{polanco2019coordinatizing}. This line of research is motivated by TDA literature extending the circular coordinate framework.

As we observed, the computational cost for computing circular coordinates
is high. One common way
of reducing the computation cost is to use sub-samples instead of
full samples in the construction of complexes \cite{otter2017roadmap}.
From the perspective of data analysis, such a sub-sampling will introduce
more uncertainty and lose some information. While we know that
sub-sampling preserves the most topological features in a dataset, it is
unclear how other (nonlinear) dimension reduction methods behave
under a sub-sampling scheme. In Section \ref{subsec:Parameter-uniform-sampling-scheme}
and Appendix \ref{subsec:Volume-uniform-sampling-scheme}, we already observed
that the reduced datasets have quite different representations
when we sample differently. This line of research aims at exploring how sub-sampling can be utilized in topological dimension reduction tasks and would be of interest for both statisticians and topologists alike.

Moreover, we know that the real coordinates in classical multidimensional scaling have an
absolute scale that depends on the particular dataset. Circular coordinates
have no absolute scale since their domain is specified to be $S^{1}$. The
circular coordinates, along with penalty functions, provide algebraically
topologically independent circular  coordinates. It will be of great
practical and theoretic interest to investigate the interaction between
algebraic independence and probabilistic independence in multidimensional
scaling \cite{de2011persistent}.

\section*{Compliance with ethical standards}
On behalf of all authors, the corresponding author states that there is no conflict of interest.
All the data we used are from published or open access articles appropriately cited in the text.
Our examples and code are available at \href{https://github.com/appliedtopology/gcc}{github.com/appliedtopology/gcc}.

\section*{Acknowledgments}
This material is based upon the work supported by the National Science Foundation under Grant No. DMS-1439786 while the authors were in residence at the Institute for Computational and Experimental Research in Mathematics in Providence, RI, during the Applied Mathematical Modeling with Topological Techniques program.

The authors would like to thank ICERM and the organizers of the ``Applied Mathematical Modeling with Topological Techniques" workshop Henry Adams, Maria D'Orsogna, Jos\'e Perea, Chad Topaz, Rachel Neville for the thought-provoking talks and a productive environment.


\appendix


\section{\textcolor{brown}{\label{sec:topol-backgr}}Topological background}

We recommend \cite{Hatcher2001} as a reference for all the details
we will be covering here. In the following discussion, we fix some
field $\mathbb{k}$. In this paper, we choose $\mathbb{k}=\mathbb{Z}_{23}=\mathbb{Z}/23\mathbb{Z}$
to be our default coefficient field $\mathbb{k}$ for computing the persistent
cohomology.

Given a set $V$ of vertices, an abstract simplicial complex is a
subset $\Sigma\subseteq2^{V}$ of the powerset, closed under subsets.
In other words, if $\tau\subset\sigma\in\Sigma$, then $\tau\in\Sigma$.
A simplex is said to have dimension $\dim\sigma=|\sigma|-1$.

To a simplicial complex, we associate a chain complex -- a sequence
of vector spaces linked by a sequence of distinguished linear maps
called the boundary maps. The chain complex $C_{*}\Sigma$ has component
vector spaces $C_{d}\Sigma$ spanned by the $d$-dimensional simplices
of $\Sigma$. The boundary map $\partial$ acts through

\[
\partial[v_{0},\dots,v_{d}]=\sum_{i}(-1)^{i}[v_{0},\dots,v_{i-1},v_{i+1},\dots,v_{n}]
\]

It is easy to show that $\partial^{2}=\partial\circ\partial=0$, and
thus $\img\partial\subseteq\ker\partial$. We define the boundaries
to be the elements in the image of $\partial$ and the cycles to be
the elements in the kernel of $\partial$.

The homology of $\Sigma$ is defined to be the quotient vector space
$H_{*}\Sigma=\ker\partial/\img\partial$. Homology can be thought
of as capturing essential or surprising cycles in $\Sigma$.

Given two maps $f,g:X\to Y$, we say that $f$ is homotopic to $g$
if there is a continuous map $H:X\times[0,1]\to Y$ such that $H(x,0) = f(x)$ and
$H(x,1) = g(x)$ for all $x\in X$. Homotopy captures the notion of two
maps being equal up to continuous deformations. Homotopy forms an
equivalence relation -- so we can talk about equivalence classes
of maps up to homotopy, or homotopy classes of maps.

Given a simplicial complex $\Sigma$, its chain complex $C_{*}\Sigma$
is the (graded) vector space spanned by the simplices in $\Sigma$,
with a boundary map $\partial:C_{*}\Sigma\to C_{*-1}\Sigma$ constructed
the usual way. The \emph{cochain comple}x of $\Sigma$, denoted by
$C^{*}\Sigma$, is the dual vector space of the chain complex: $C^{i}\Sigma=\Hom(C_{i}\Sigma,\mathbb{k})$
with a \emph{coboundary operator} $\delta:C^{*}\Sigma\to C^{*+1}\Sigma$
defined through $\delta f=f\circ\partial$. The \emph{cohomology}
$H^{*}\Sigma$ of $\Sigma$ is the homology of the cochain complex
of $\Sigma$. We denote by \emph{coclass} the elements of cohomology;
\emph{cocycle} the elements of the kernel of the coboundary map; \emph{coboundary}
the elements of the image of the coboundary map.

Using persistent cohomology, \cite{de2011persistent} shows that
high-dimensional nonlinear data can be represented in the form of
low-dimensional circular coordinates. Cohomologies over the integers
have especially attractive properties for our work: $H^{1}(\Sigma,\mathbb{Z})$
is in bijective correspondence with homotopy classes of maps $\Sigma\to S^{1}$.
The correspondence is constructive: if $[f]\in H^{1}\Sigma$, then
$f(u,v)\in\mathbb{Z}$ for any edge $[uv]$ in the complex. We construct
a map onto the circle by mapping all vertices to a single point on
the circle, and by mapping an edge $[uv]$ to wrap around the circle
as many times as $f(u,v)$ specifies. In \cite{de2011persistent},
the authors describe how to go from such a map to one that smoothly
spreads the vertex images around the circle using least squares optimization:
$[f]\in H^{1}(\Sigma,\mathbb{Z})$ is interpreted as an element of
$H^{1}(\Sigma,\mathbb{R})$. Moreover, the modular reduction $z\mod{1.0}$
is a function from the vertices of $\Sigma$ to the circle with the
required smoothness properties.

 \section{\label{sec:PCA-pre-processed-pipeline_gpca}Additional comparison of the circular coordinates to GPCA}

Continuing the comparison of the circular coordinates and PCA in Section
\ref{sec:PCA-pre-processed-pipeline}, we present the analysis of
the GPCA presentation \cite{VidalMS2005} of the same dataset $X\subset\mathbb{R}^{3}$
in Section \ref{sec:PCA-pre-processed-pipeline}. GPCA generalizes PCA and can fit data lying in a union of subspaces. This is done by representing a union of subspaces with a set of homogeneous polynomials on the covariates, and then running PCA on these polynomials.

\begin{figure}[t!]
\centering

\begin{subfigure}{0.45\linewidth}\centering\includegraphics[height=5.3cm]{figs/fig_cc-pca_dataset_scatter}\caption{Scatter plot of $X$.}
\label{fig:cc-pca_gpca_dataset_scatter}\end{subfigure} \hspace{5mm}
\begin{subfigure}{0.45\linewidth}\centering\includegraphics[height=5.3cm]{figs/fig_cc-pca_dataset_ph}\caption{The $1$-dimensional persistent cohomology of $X$.}
\label{fig:cc-pca_gpca_dataset_ph}\end{subfigure}

\begin{subfigure}{0.45\linewidth}\centering\includegraphics[height=5.3cm]{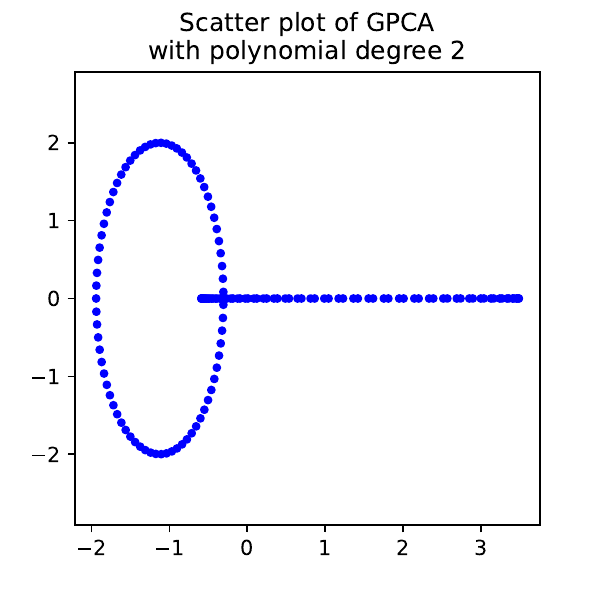}\caption{Scatter plot of $X^{gpca,2}$.}
\label{fig:cc-pca_gpca2_scatter}\end{subfigure} \hspace{5mm} \begin{subfigure}{0.45\linewidth}\centering\includegraphics[height=5.3cm]{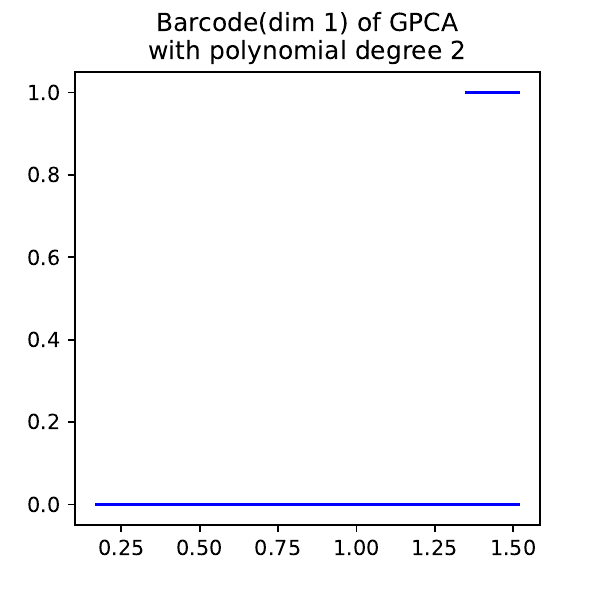}\caption{The $1$-dimensional persistent cohomology of $X^{gpca,2}$.}
\label{fig:cc-pca_gpca2_ph}\end{subfigure}

\begin{subfigure}{0.45\linewidth}\centering\includegraphics[height=5.3cm]{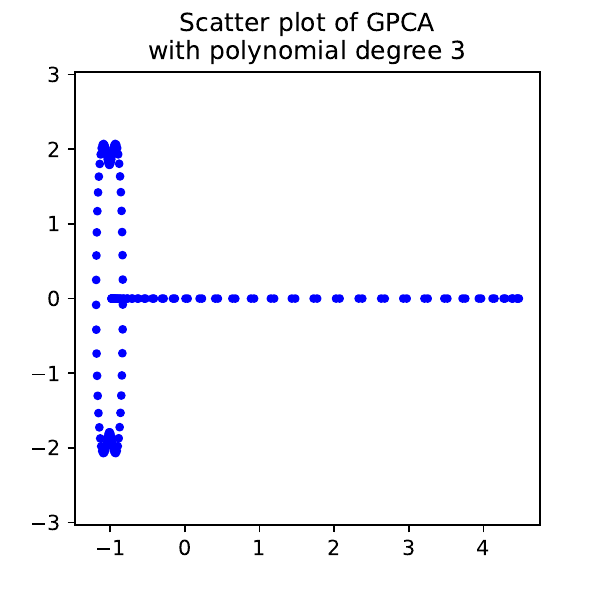}\caption{Scatter plot of $X^{gpca,3}$.}
\label{fig:cc-pca_gpca3_scatter}\end{subfigure} \hspace{5mm} \begin{subfigure}{0.45\linewidth}\centering\includegraphics[height=5.3cm]{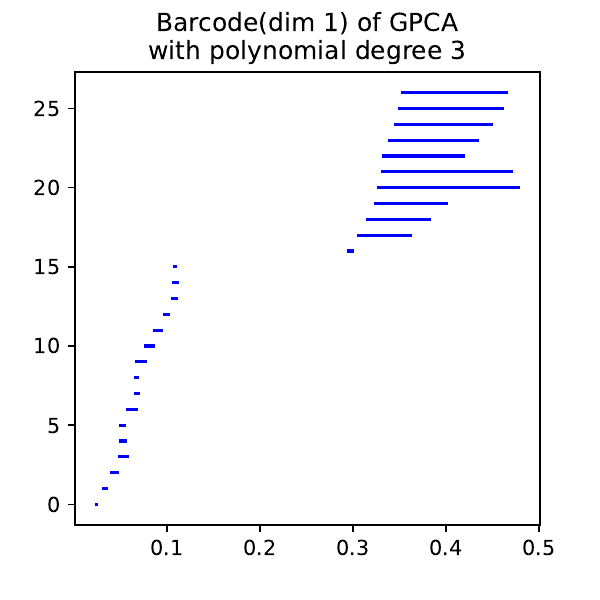}\caption{The $1$-dimensional persistent cohomology of $X^{gpca,3}$.}
\label{fig:cc-pca_gpca3_ph}\end{subfigure}

\caption{The GPCA representation $X^{gpca,2}$ and $X^{gpca,3}$ of the embeddings from the first $2$ principal
components of the homogeneous polynomials of degree $2$ and $3$, respectively.}

\label{fig:cc-pca_gpca}
\end{figure}

For applying the GPCA to our data, we choose to consider the embedding of the first 2 principal
components for comparison with PCA and circular coordinates. Let $X^{gpca,2}$
be the GPCA representation of the dataset $X$ with the homogeneous polynomials of degree $2$, and $X^{gpca,3}$ be the GPCA representation with the homogeneous polynomials of degree $3$. From Figure \ref{fig:cc-pca_gpca}\subref{fig:cc-pca_gpca2_scatter},
we can see that one of the $1$-dimensional coboundaries of the original
data $X$ is collapsed and the $1$-dimensional cohomology structure
of $X$ is distorted in both $X^{gpca,2}$ and $X^{gpca,3}$. And
the collapsed $1$-dimensional cohomology structure is also unidentifiable
using persistent cohomology of the embedded dataset $X^{gpca,2}$
and $X^{gpca,3}$ as seen in Figure~\ref{fig:cc-pca_pca}\subref{fig:cc-pca_gpca2_ph}
and \subref{fig:cc-pca_gpca3_ph}.

\FloatBarrier

\section{\label{sec:Comparison-with-other}Comparison with other dimension
reduction methods}
There are two common approaches to analyze the reduced dataset further $\Theta(X)\subset\mathbb{T}^{k}$
within statistical framework. One way is to modify the statistical procedure so that they can take the torus $\mathbb{T}^{k}$
as the input space and its geodesic distance as the endowed metric on it. The other way is to apply an embedding map $\imath:\mathbb{T}^{k}\hookrightarrow\mathbb{R}^{l}$
to the dimension reduced data $\Theta(X)$ so that $\imath(\Theta(X))$
now lies on another Euclidean space $\mathbb{R}^{l}$. For example,
by applying an embedding map $\imath:\mathbb{T}^{k}\hookrightarrow\mathbb{R}^{2k}$
which sends each coordinate to $S^{1}$ in $\mathbb{R}^{2}$, the
dimension reduced data $\Theta(X)$ is embedded as $\imath(\Theta(X))$,
which now lives in $\mathbb{R}^{2k}$. For a better comparison with
other embedding methods, we will focus on the latter approach throughout
the paper. For visualization purposes, we specify the retained dimension to be 2 in this comparison. We retain the first two 1-cocycles with the longest persistence as the reduced coordinates.

For this high-dimensional congress voting
dataset, we provide both qualitative and quantitative analysis of circular
coordinate representations along with other NLDR methods. We observe
that circular coordinate representation seems to preserve the clusterings
well, compared to other NLDR methods. On one hand, circular coordinates
do not transform the original dataset but only add circular coordinate
values from each significant 1-cocycle for each point in the dataset.
This prevents the loss of information in the procedure of dimension
reduction. On the other hand, circular coordinates also provide local
information in terms of sub-coordinates with respect to different
significant 1-cocycles. It allows us to examine the local information
closely through visualization. In practice, it is not always true
that significant 1-cocycles arise naturally. Empirically, in high-dimensional
datasets, there are plenty of nontrivial circular structures. When significant
1-cocycles do not exist, our method cannot be applied. Hence, we supplement
our analysis with other dimension reduction methods for comparison.
However, we remark that when no significant 1-cocycle exists, it is
more beneficial to utilize the NLDR methods like t-SNE, UMAP and Laplacian
eigenmap or even linear dimension reduction in specific applications.

\begin{figure}[t!]
\centering
\begin{adjustbox}{center}
\includegraphics[scale=0.35]{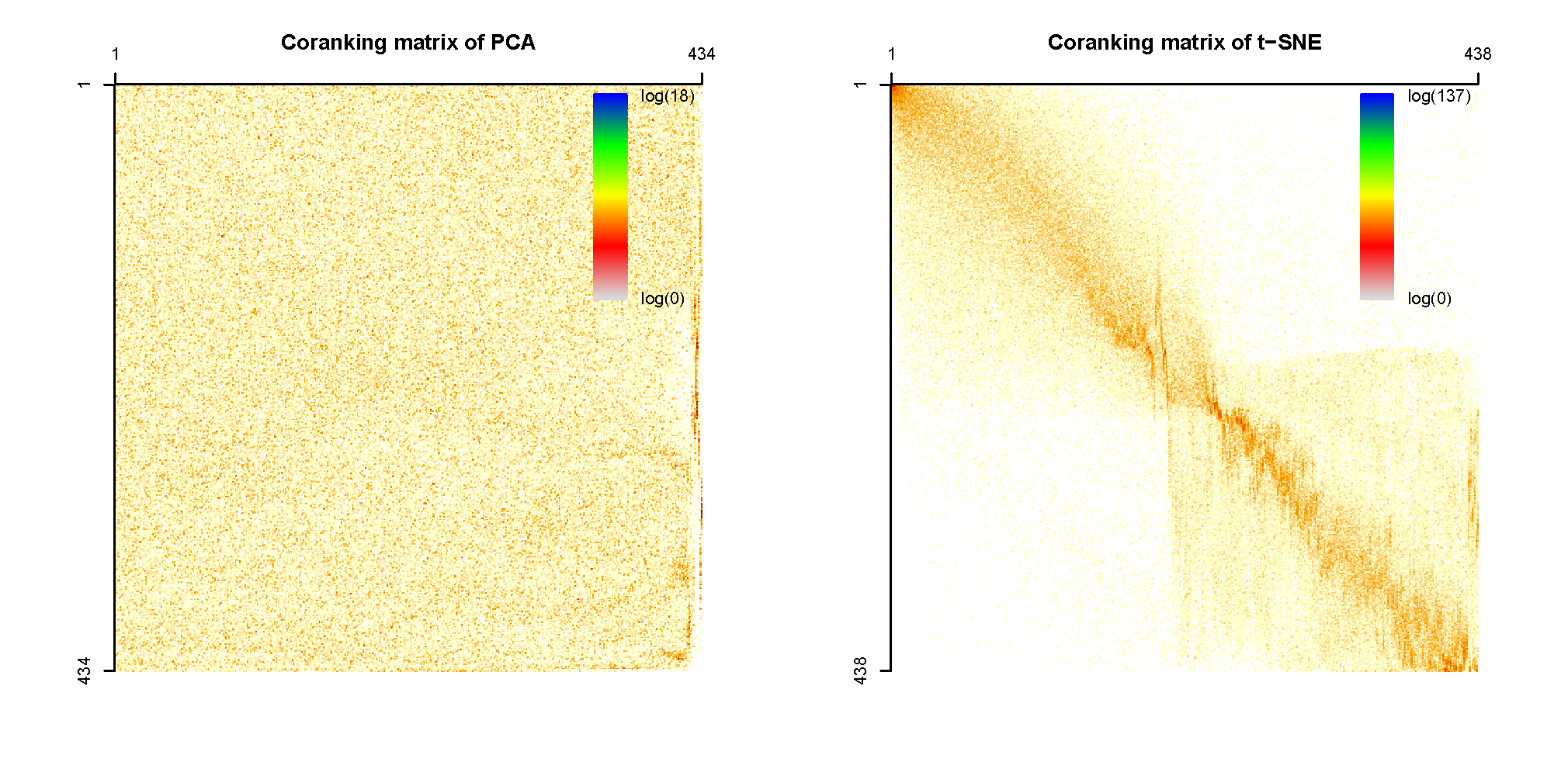}
\end{adjustbox}
\\
\begin{adjustbox}{center}
\includegraphics[scale=0.35]{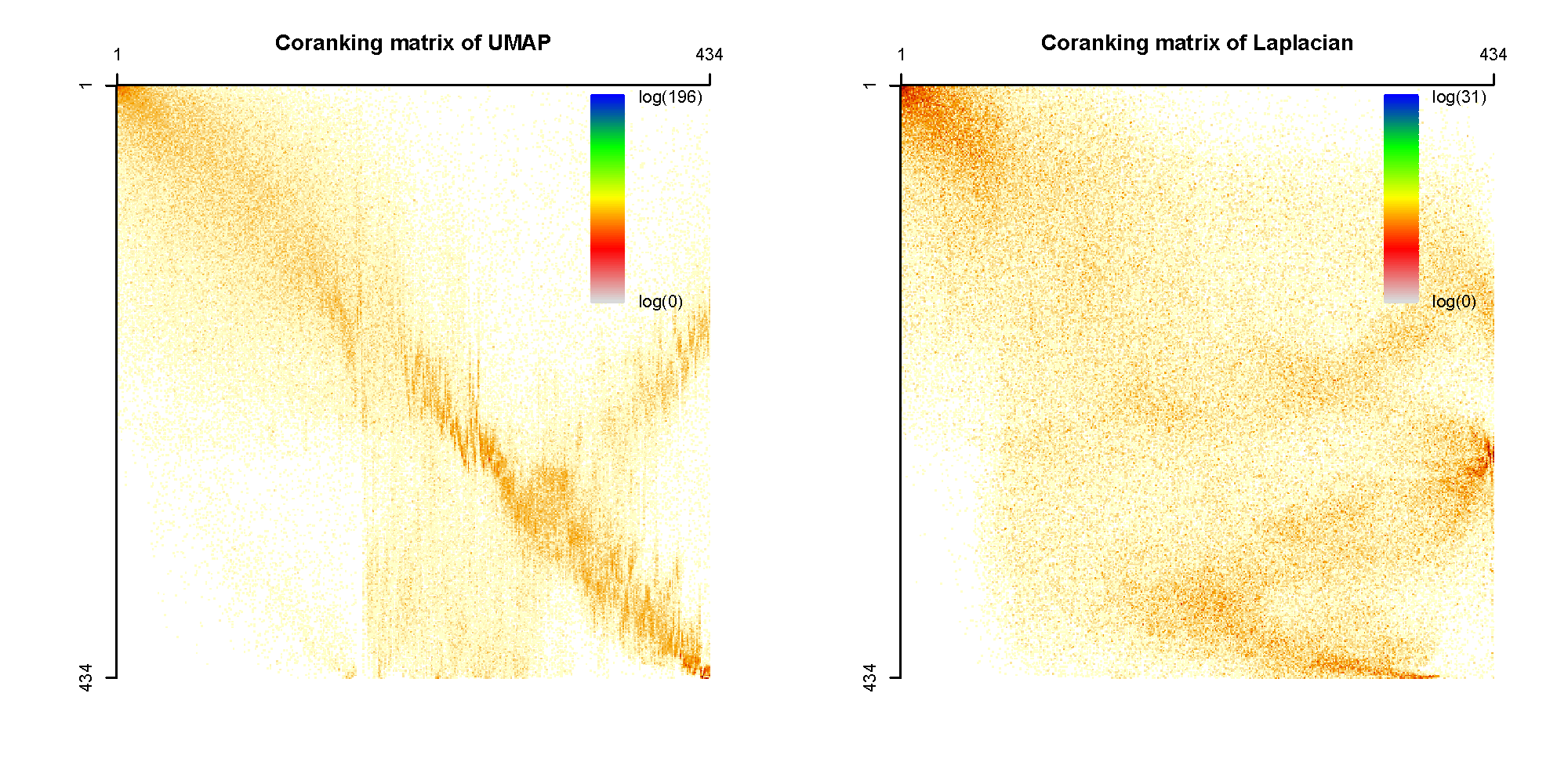}
\end{adjustbox}
\\
\begin{adjustbox}{center}
 \includegraphics[scale=0.35]{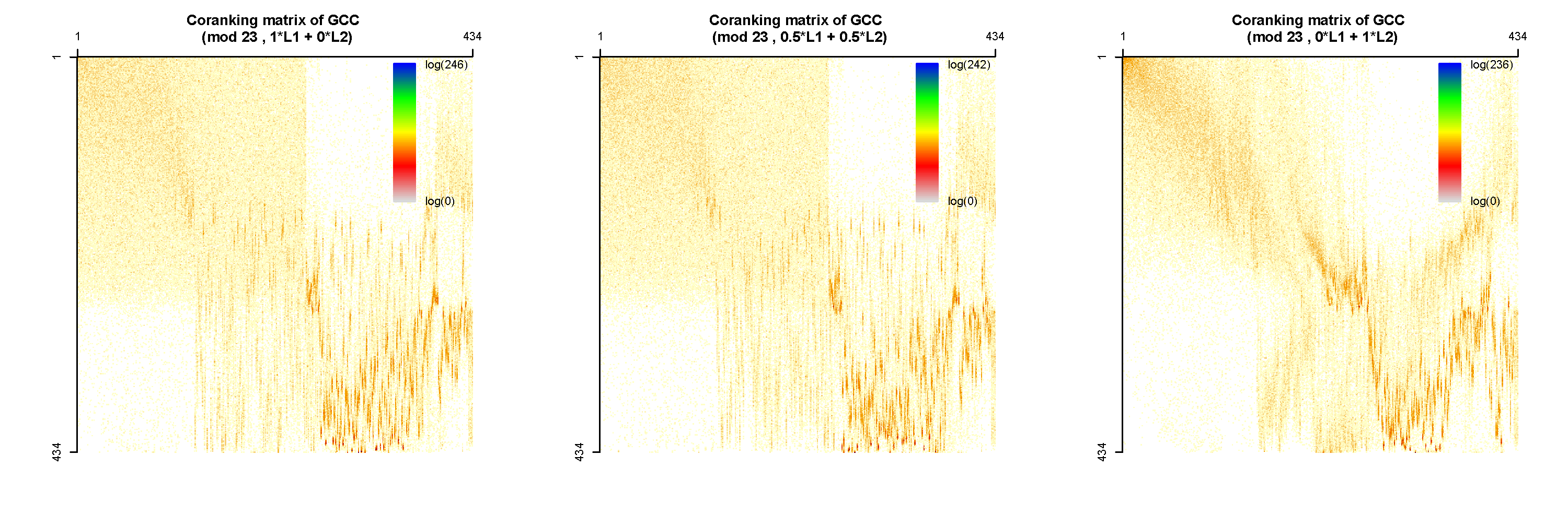}
\end{adjustbox}
\caption{\label{fig:Example Compare QC} Evaluation of dimension reduction
results obtained from different NLDR methods with the congress voting
dataset of year 1990. We display the coranking matrices
of PCA and t-SNE in the first row, and the coranking matrices of UMAP and Laplacian eigenmap in the second row. We display
the coranking matrices of circular coordinates with penalty functions
$L_{1}$, elastic norm, and $L_{2}$ in the third row.
}
\end{figure}

For PCA method, we use the $\mathtt{prcomp}$ provided in $\mathtt{R-base}$;
for t-SNE method, we use the $\mathtt{Rtsne}$ package \cite{Rtsne};
for UMAP we use the $\mathtt{umap}$ wrapper package \cite{mcinnes2018umap};
for Laplacian eigenmap we use the $\mathtt{dimRed}$ package \cite{RJ-2018-039}.
Quantitative comparison of dimension reduction methods is an ongoing
research topic \cite{lee2009quality,gupta2011evaluating,lueks2011evaluate}.
Since most of the quantitative measures are based on the coranking matrix
\cite{lee2009quality,lueks2011evaluate}, we provide plots of
coranking matrices of the dimension reduced results computed from
$\mathtt{coRanking}$ package \cite{RJ-2018-039} in Figure \ref{fig:Example Compare QC}.
Compared to PCA, nonlinear dimension reduction methods clearly retains
the corank information better. Among NLDR methods under consideration,
the t-SNE and UMAP have matrix structures more concentrated around
diagonal than Laplacian eigenmap, and they preserve the clustering
across parties well. Unexpectedly, the Laplacian eigenmap cannot preserve
the party clustering well even with nonlinear construction. It is
not hard to see that the block structure in the circular coordinate
representation is clearly exhibited, which is an indication that circular
coordinate representation retains information and has fewer hard intrusions/extrusions
\cite{lueks2011evaluate}. Compared
with the $L_{2}$ penalty, the circular coordinates with generalized
penalty have even sharper block structures.

\begin{figure}[t]
\centering
\begin{adjustbox}{center}
\includegraphics[scale=0.35]{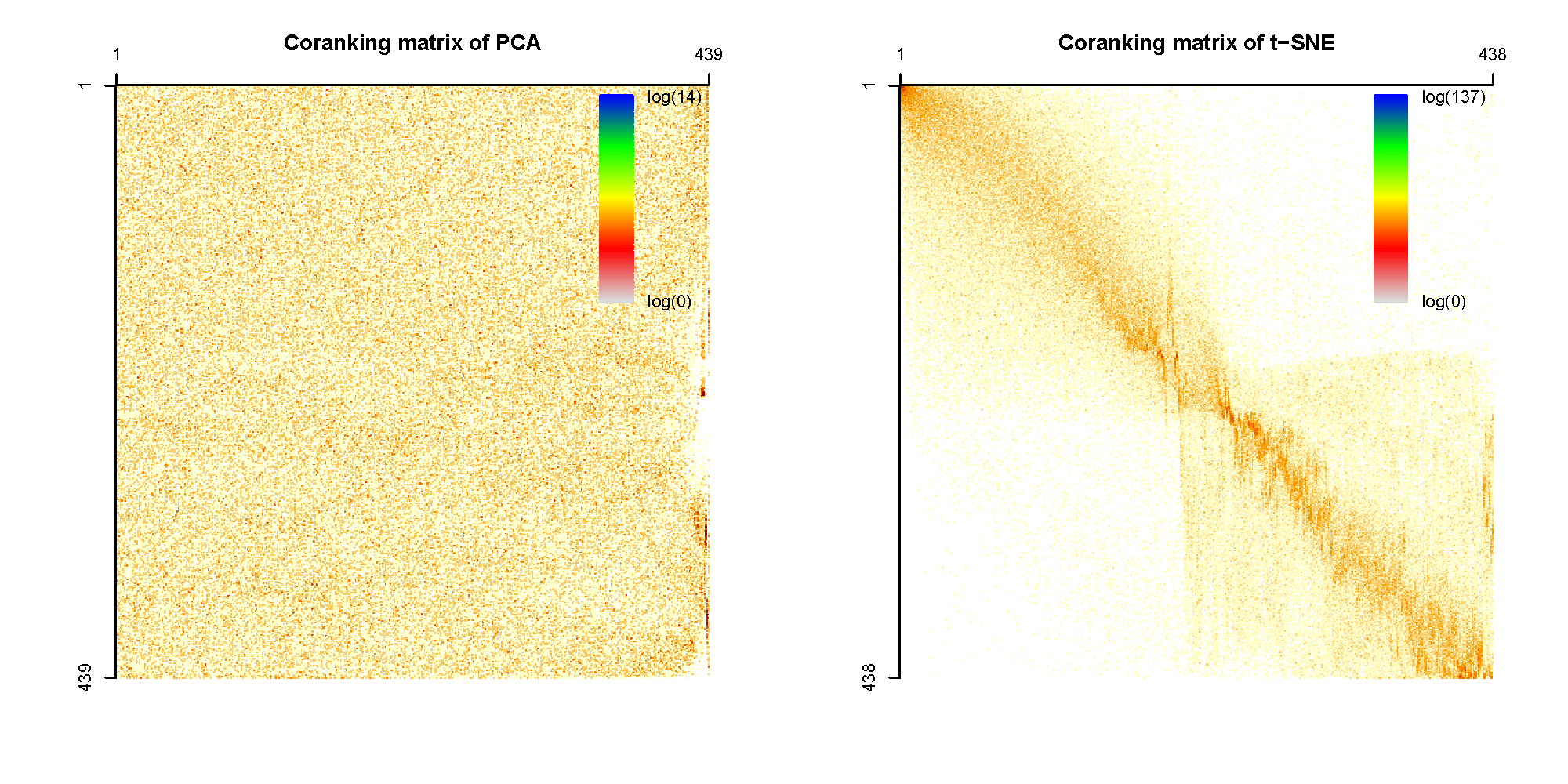}
\end{adjustbox}
\\
\begin{adjustbox}{center}
\includegraphics[scale=0.35]{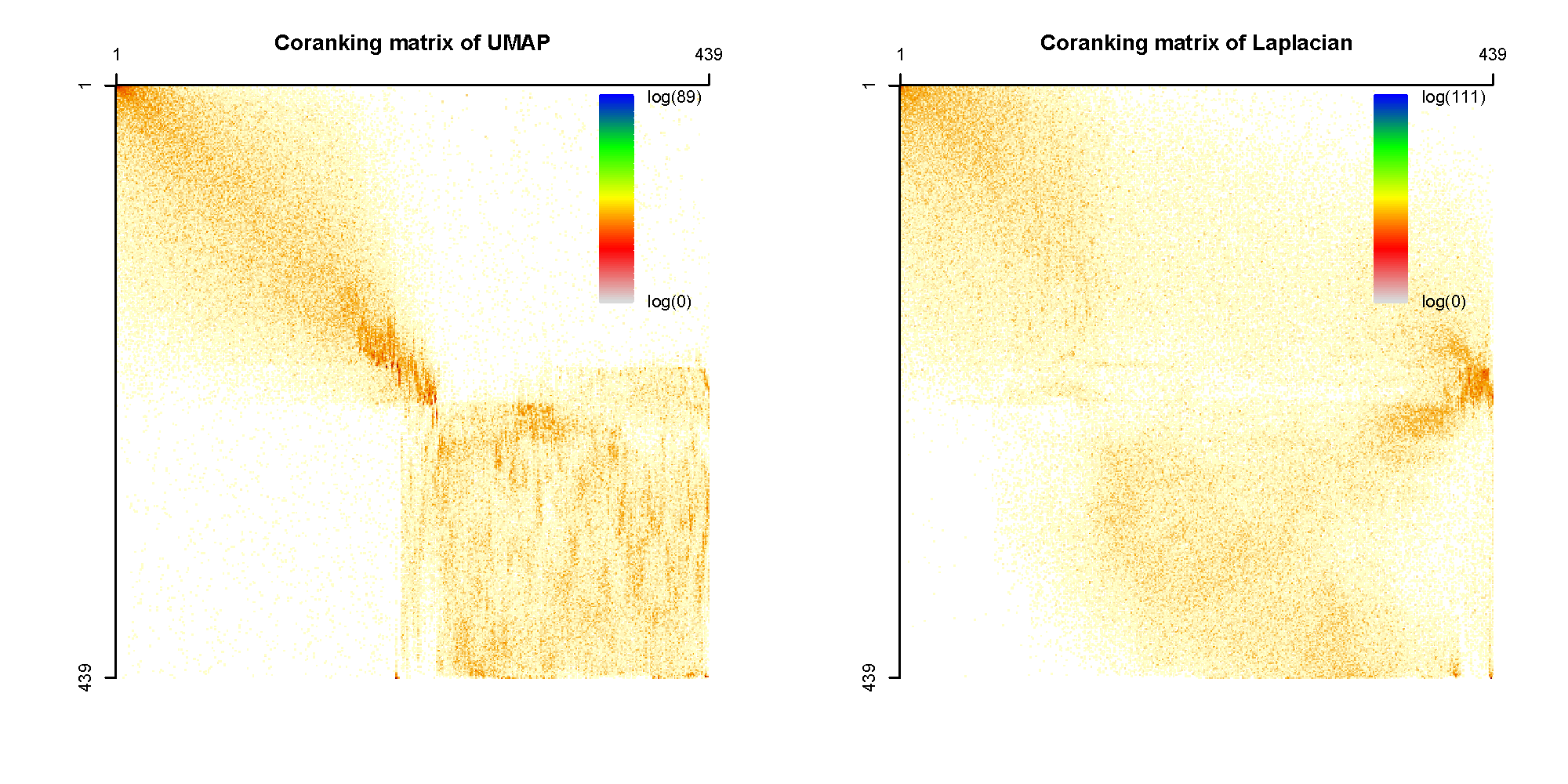}
\end{adjustbox}
\\
\begin{adjustbox}{center}
 \includegraphics[scale=0.35]{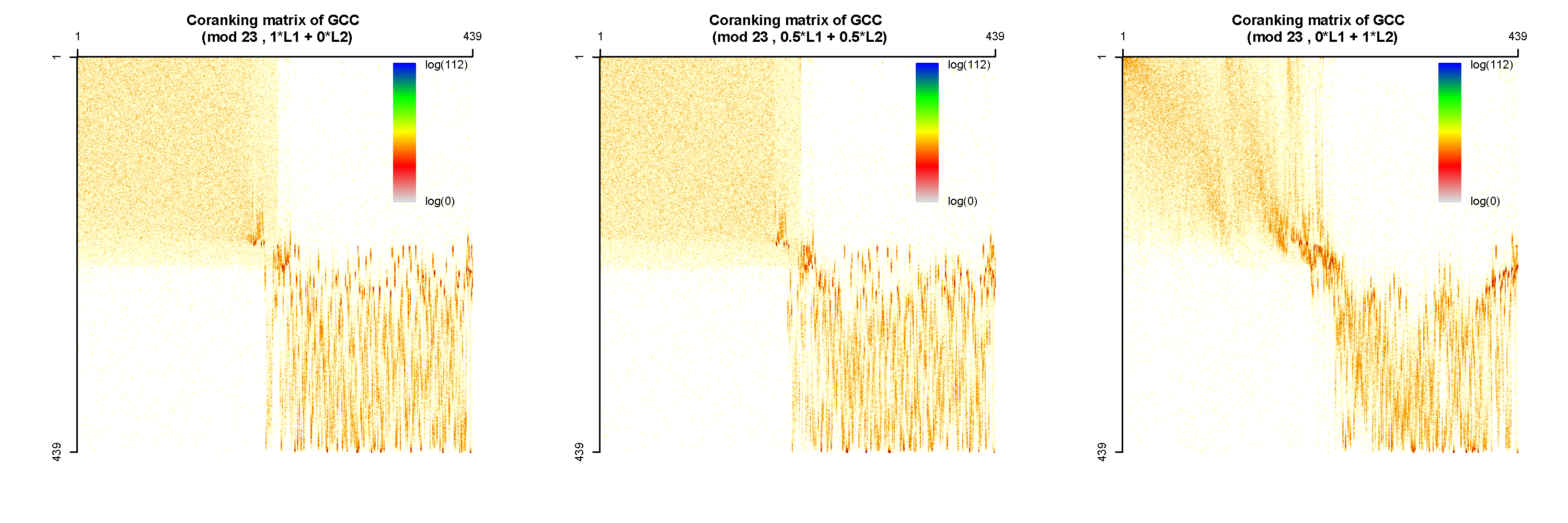}
\end{adjustbox}
\caption{\label{fig:Example Compare QC-1} Evaluation of dimension reduction
results obtained from different NLDR methods with the congress voting
dataset of year 1998. We display the coranking matrices
of PCA and t-SNE in the first row, and the coranking matrices of UMAP and Laplacian eigenmap in the second row. We display
the coranking matrices of circular coordinates with penalty functions
$L_{1}$, elastic norm, and $L_{2}$ in the third row.
}\bigskip
\end{figure}

\begin{figure}[t!]
\begin{adjustbox}{center}

\includegraphics[scale=0.35]{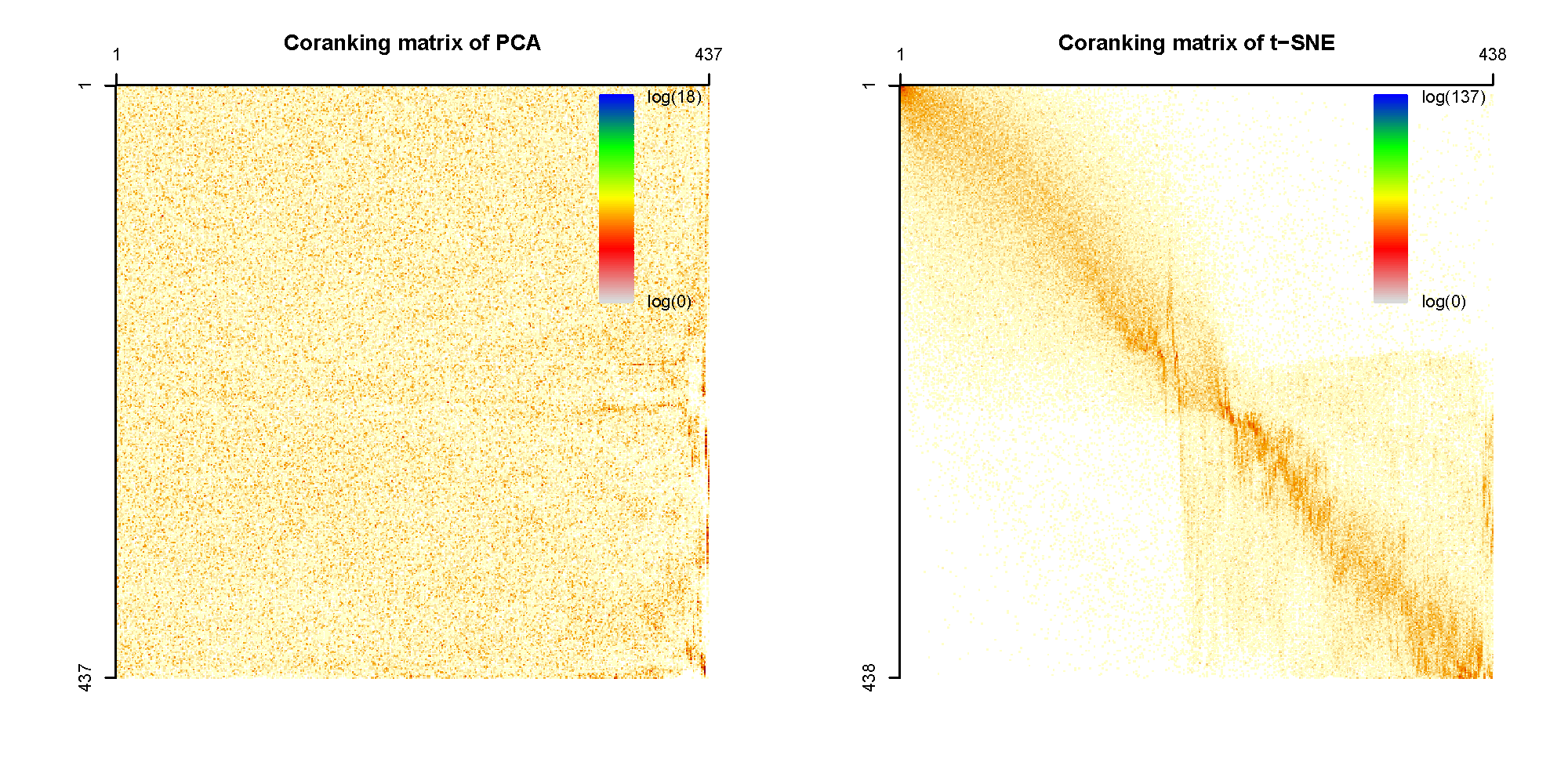}
\end{adjustbox}
\\
\begin{adjustbox}{center}
\includegraphics[scale=0.35]{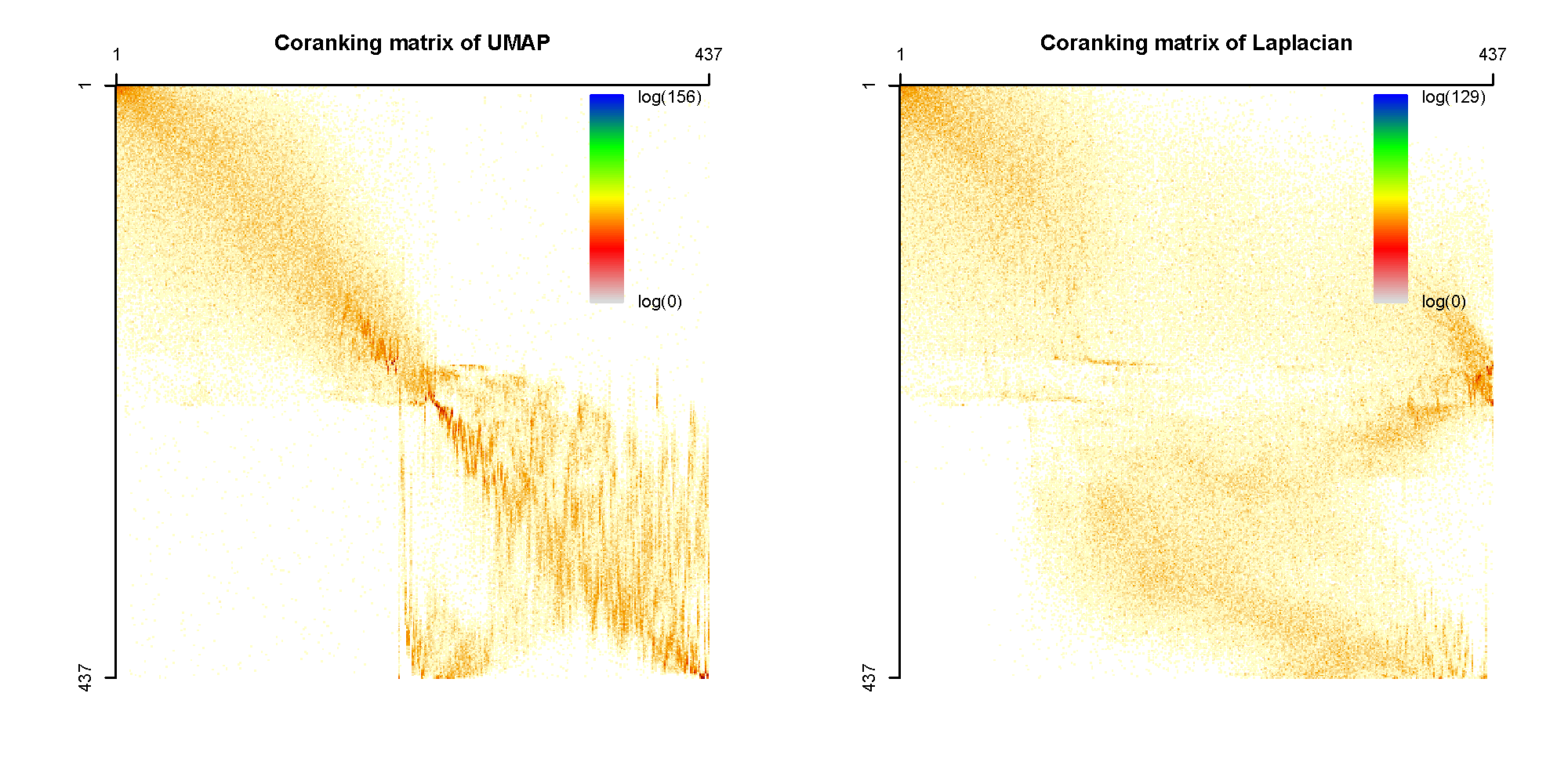}
\end{adjustbox}
\\
\begin{adjustbox}{center}
 \includegraphics[scale=0.35]{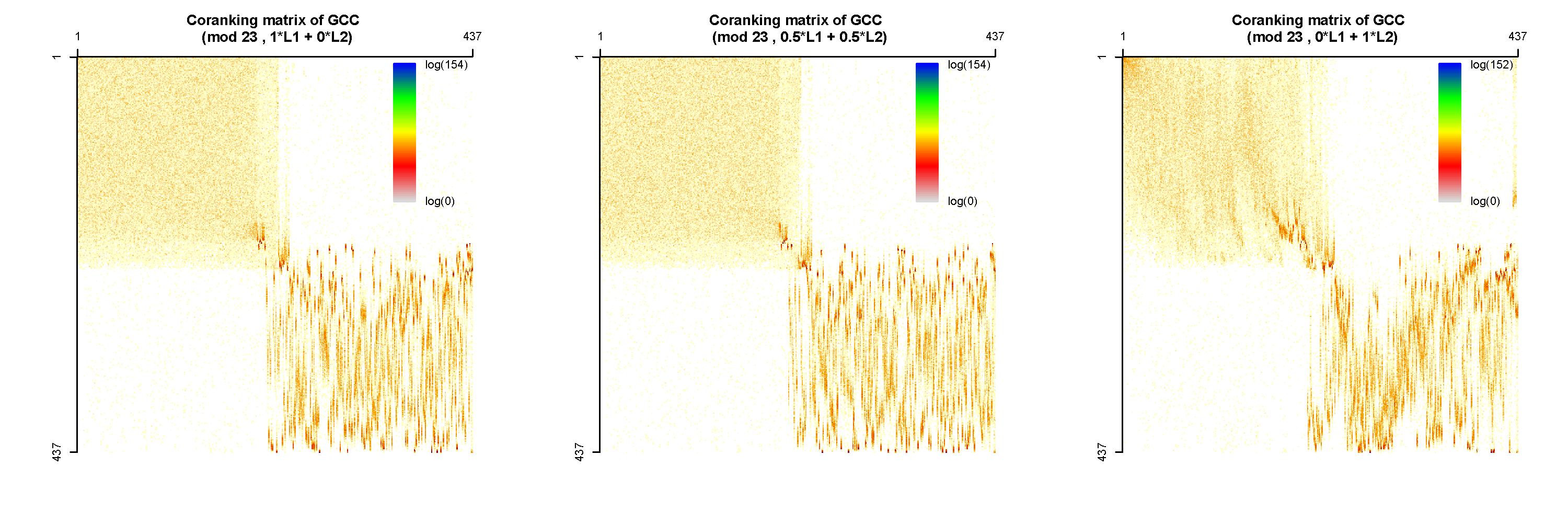}
\end{adjustbox}

\caption{\label{fig:Example Compare QC-2} Evaluation of dimension reduction
results obtained from different NLDR methods with the congress voting
dataset of year 2006. We display the coranking matrices
of PCA and t-SNE in the first row, and the coranking matrices of UMAP and Laplacian eigenmap in the second row. We display
the coranking matrices of circular coordinates with penalty functions
$L_{1}$, elastic norm and $L_{2}$ in the third row.
}\bigskip
\end{figure}
\FloatBarrier

\section{\label{subsec:Volume-uniform-sampling-scheme}Volume uniform sampling scheme}

\cite{niyogi2008finding} pointed out that we need sufficient samples to recover the homology
type of manifold; \cite{tausz2011applications} (Section 3.3) also remarked that the sample size is important in uniform sampling. Although it may be more straightforward to use a
uniform sampling on the parameter space when we try to draw random samples from its underlying manifold $M$, doing so may cause an insufficient sampling
of the manifold surface.

Such a ``volume non-uniform'' sample is caused by a sampling scheme not proportional to the volume form of the manifold. It would not be the best descriptor of the support of the distribution.

Our discussion below points out that the choice of the sampling scheme, even though it does affect the circular coordinate representation, it is unlikely that it would qualitatively change the representation.

Let us consider again the two-dimensional disc $\{(x,y)\in\mathbb{R}^{2}\mid x^{2}+y^{2}\leq1\}$
parameterized by $\Phi:(r,\theta)\mapsto(r\cos\theta,r\sin\theta)\in\mathbb{R}^{2}$.
We now consider its Jacobian (or volume element) $\left|\frac{\partial\Phi}{\partial(r,\theta)}\right|=r$
with maximum $1$. To build a \emph{volume uniform sampling scheme}, we first
sample pairs uniformly on $[0,1]\times[0,2\pi]$. However, every time
we sample such a pair, we independently sample another random variable
$u\sim\Uniform(0,1)$. We only retain such a pair $(r,\theta)$ and
hence a point $(r\cos\theta,r\sin\theta)$ only if $u\leq r$, the
Jacobian evaluated at $(r,\theta)$. This rejection sampling scheme
is not uniform over the parameter space but it samples proportional
to the volume element of the disc. Again, we choose the most significant
cocycle and obtain the circular coordinates for this dataset. 

\begin{figure}[t!]
\centering

\begin{tabular}{cc}
circular coordinates & correlation plot\tabularnewline
\hline
\includegraphics[height=5.7cm,page=10]{figs/Example1_J_output_edit.pdf} &
\includegraphics[height=5.7cm]{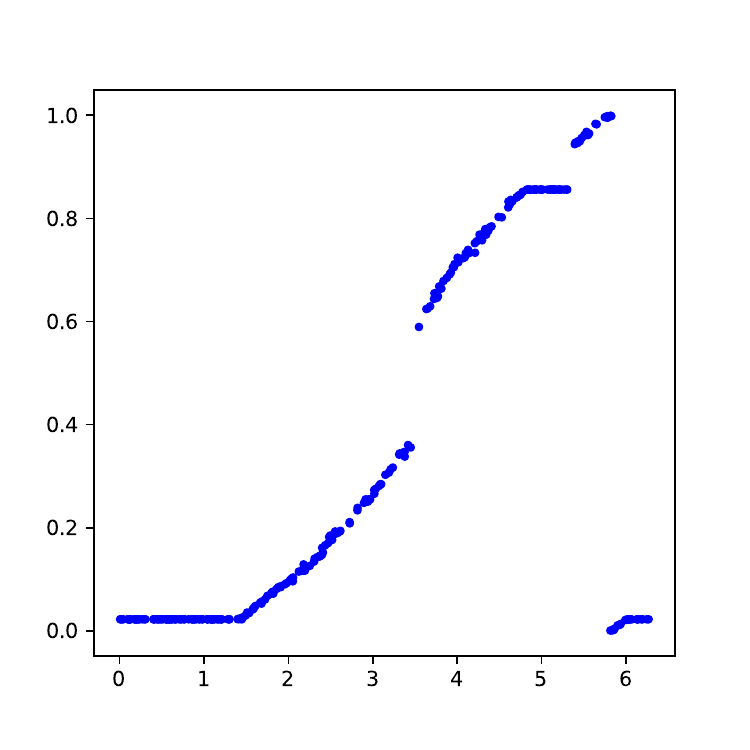}
\tabularnewline
\includegraphics[height=5.7cm,page=17]{figs/Example1_J_output_edit.pdf} &
\includegraphics[height=5.7cm]{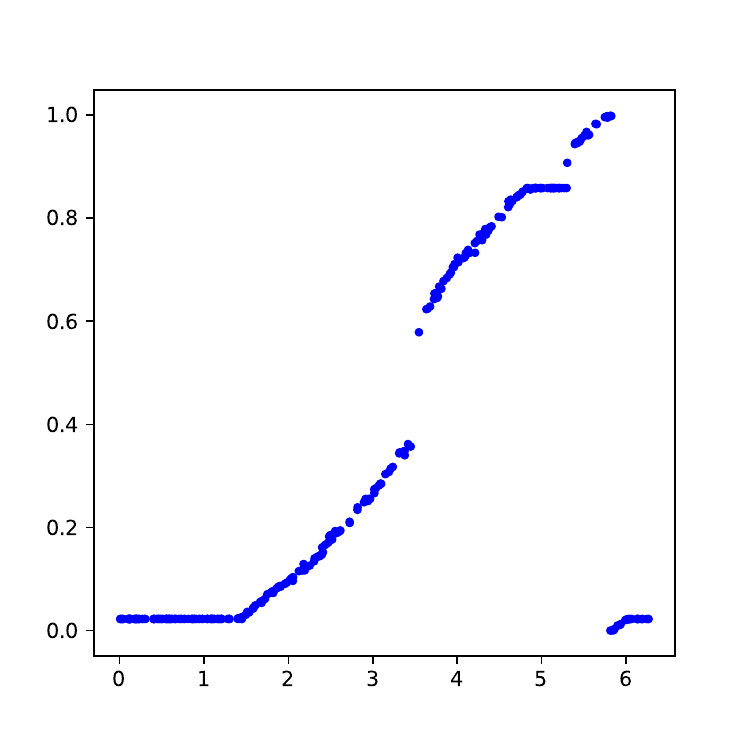}
\tabularnewline
\includegraphics[height=5.7cm,page=24]{figs/Example1_J_output_edit.pdf} &
\includegraphics[height=5.7cm]{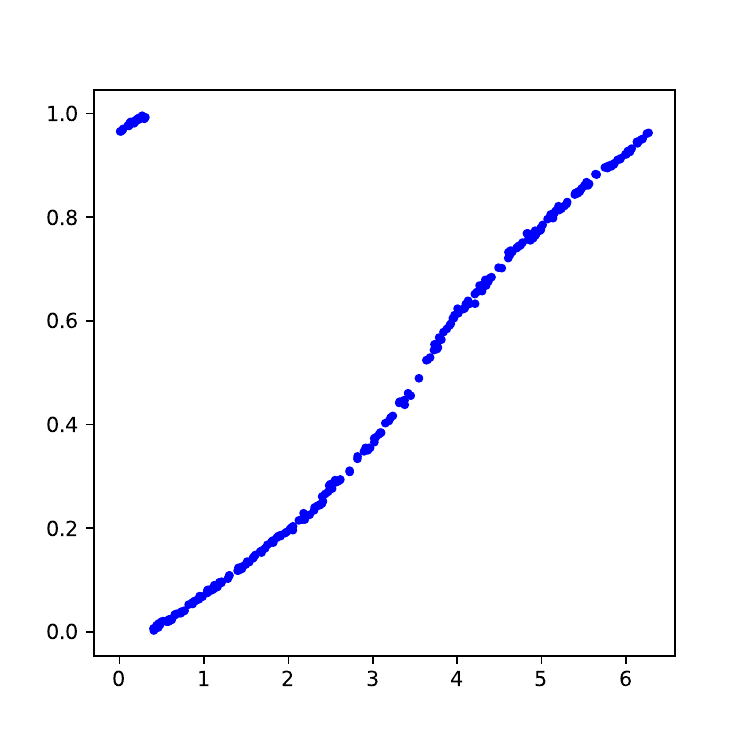}
\tabularnewline
\end{tabular}
\caption{\label{fig:Example 5}Example 5: The $L_{2}$ smoothed and generalized
penalized circular coordinates of the Jacobian rejection sampled dataset
($n=300$) from a ring with fixed width (Jacobian rejection sampling).
The first, second, and the third row correspond to $\lambda=0$, $0.5$,
and $1$, respectively.}
\end{figure}

The essence of the volume uniform sampling scheme is that it samples from the manifold $M$ according to the distribution of the area (or volume) element. And it will not lead to a parameter uniform sampling scheme in the parameter space, but the rejection sampling based on Jacobian will sample proportional to the first fundamental form, or the volume element of the manifold to ensure that the density of sample points is evenly distributed. Therefore, we call this sampling scheme the \emph{volume uniform sampling}.
Even a visual comparison between Figure
\ref{fig:Example-1:-Ring} and Figure \ref{fig:Example 5} provides strong
evidence that these two sampling schemes are completely different in the distribution of the sample points over the same ring,
especially near the boundaries of the ring.

\begin{figure}[t!]
\centering

\begin{tabular}{cc}
circular coordinates & correlation plot\tabularnewline
\hline
\includegraphics[height=5.7cm,page=10]{figs/Example_pinched__torus_J_output_edit.pdf} &
\includegraphics[height=5.7cm]{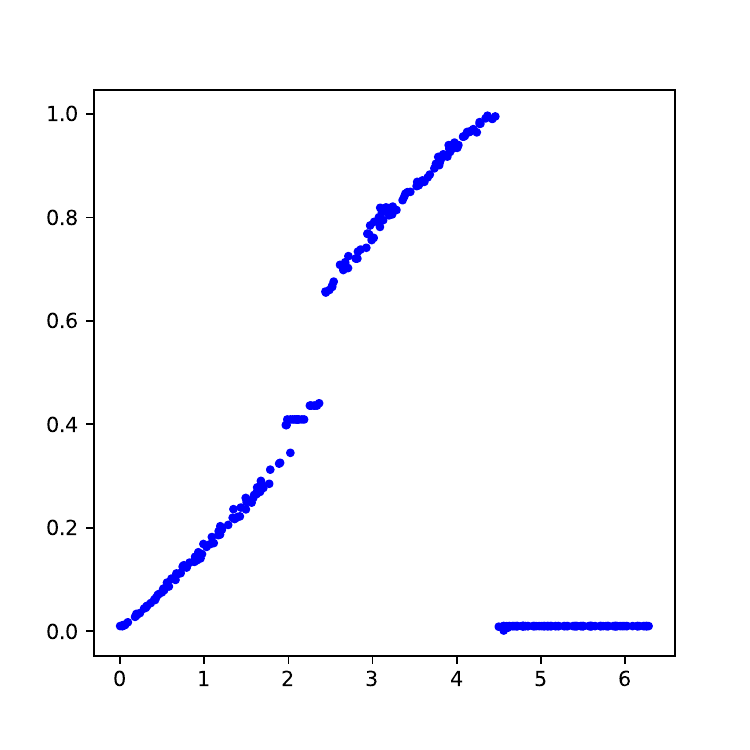}
\tabularnewline
\includegraphics[height=5.7cm,page=17]{figs/Example_pinched__torus_J_output_edit.pdf} &
\includegraphics[height=5.7cm]{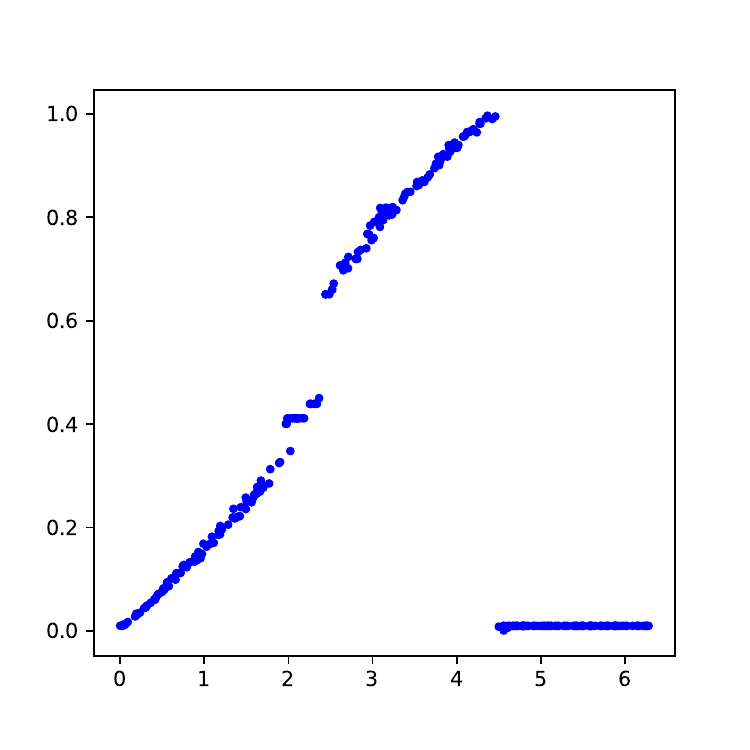}
\tabularnewline
\includegraphics[height=5.7cm,page=24]{figs/Example_pinched__torus_J_output_edit.pdf} &
\includegraphics[height=5.7cm]{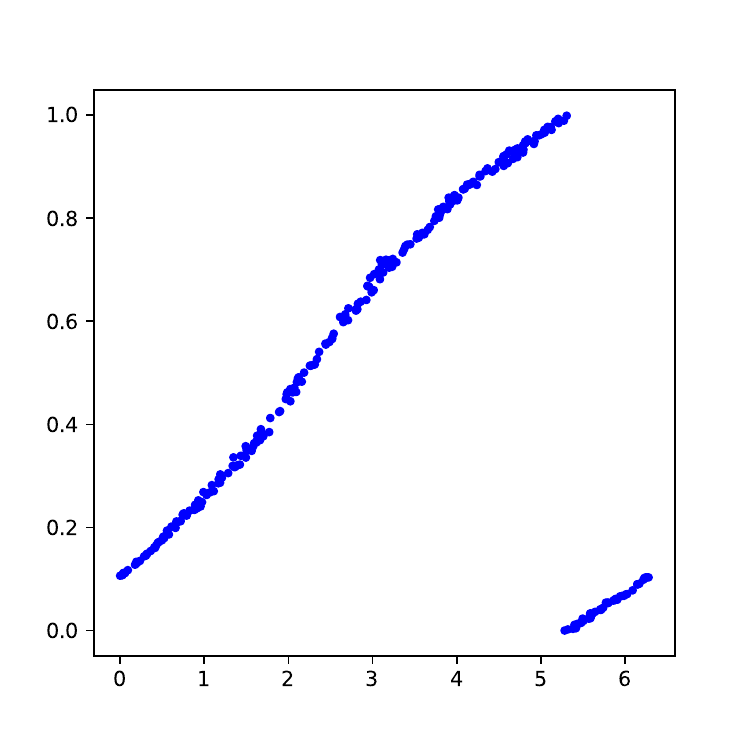}
\tabularnewline
\end{tabular}

\caption{\label{fig:Example 6}Example 6: The $L_{2}$ smoothed and generalized
penalized circular coordinates of the Jacobian rejection sampled dataset
($n=300$) from a Dupin cyclide with $r=2$, $R=1.5$ as in Section \ref{subsec:Example-3:-Dupin}.
The first, second, and the third row correspond to $\lambda=0$, $0.5$,
and 1, respectively.}
\end{figure}

We use the Jacobian for our rejection sampling in Figure \ref{fig:Example 5} because it is a 2-dimensional object in $\mathbb{R}^2$ and the Jacobian serves as a volume element. When the manifold is in a general position, we want to consider the volume element as rejection criterion. As shown in Figure \ref{fig:Example 6}, we can compute the circular coordinates obtained from rejection sampling on the Dupin cyclide with $r=2$, $R=1.5$ as we investigated in Section \ref{subsec:Example-3:-Dupin}. Dupin cyclide is a surface in $\mathbb{R}^3$, therefore, the rejection sampling is based on the volume element given by its first fundamental form.  When comparing Figure \ref{fig:Example-3} and \ref{fig:Example 6}, the difference becomes even more obvious. Since the volume uniform sampling scheme forces the sample points to occupy the space and spread out more evenly.

Similar contrasts between parameter uniform and volume uniform samplings could be observed for the spherical shells in $\mathbb{R}^3$ and $\mathbb{R}^4$.
From the simulation results in Figures \ref{fig:Example 5} and \ref{fig:Example 6}, we observe that the sampling scheme on $M$ is by far the strongest factor that affects the distribution of the constant edges. In regions with high sampling density, the changes of coordinate values are highly penalized by $L_1$ or generalized penalty functions in the cohomolougous optimization problem (\ref{eq:cohomologous opt - L1}).

On one hand, the choice
of penalty functions will also exhibit different levels of sensitivity
for different sampling schemes. $L_2$ penalty does not generate many constant edges but still shows a larger region of data points with coordinate values without much variation. Generalized penalty functions generate a lot of constant edges. Unlike the uniformly sampled case in Figure \ref{fig:Example-1:-Ring}, there are fewer clusters of constant edges. However, neither $L_2$ nor the generalized penalty produces a qualitative difference in the distribution of constant edges.

On the other hand, although we did not observe a qualitative difference between the color scale visualization of circular coordinates in Figure \ref{fig:Example-1:-Ring} and \ref{fig:Example 5}, we can clearly see that under different sampling schemes, the (angle) correlation plots are somehow different and so is the concentration of constant edges. While under the $L_2$ penalty the correlation plot shows a difference in slopes, $L_1$ penalty and the elastic norm also produces a  difference in coordinate values. There are two different dotted lines in the first and second rows in Figure \ref{fig:Example-1:-Ring} but only one dotted line in the corresponding rows of Figure \ref{fig:Example 5}. We pointed out here that the sampling scheme, although it does not affect the qualitative feature detection (i.e., distribution and denseness of constant edges), it will affect coordinate values and can be spotted from the correlation plot associated with the coordinates. This is true no matter if we are using $L_2$ or generalized penalty functions. 

In this way, circular coordinates may as well provide an informative reference when we are interested in the sampling scheme on high-dimensional datasets. In short, under different sampling schemes, the circular coordinates under $L_2$ penalty functions:\label{properties2}
\begin{enumerate}
\item Would not create a qualitative difference in the distribution of coordinate values, and hence the distribution of constant edges.
\item Would usually display significant difference in the correlation plots associated with the circular coordinates.
\end{enumerate}
In contrast, since the circular coordinates under $L_1$ and generalized penalty functions accommodate the sparsity in the dataset, when the rejection sampling distributes the sample points more evenly over the manifold, we would observe that the circular coordinates with a  generalized penalty function under a difference sampling scheme:\label{properties3}
\begin{enumerate}
\item Would generate a qualitative difference in the distribution of constant edges.
\item Would also display a difference in the correlation plots associated with the circular coordinates.
\end{enumerate}
The observations we obtained from the comparison between parameter uniform and volume uniform schemes in this section provide additional evidence to the claim that the sampling scheme of the dataset is an important factor in TDA \cite{niyogi2008finding,tausz2011applications}. It also brings up a new question that how the topology in the approximating complex $\Sigma$ could reflect the empirical distribution. In asymptotics, when the sample size $n\rightarrow\infty$, we expect that the approximating complex $\Sigma$ would have the same topology as $M$; and we also expect that the empirical distribution would converge to the true density. Therefore, it remains an interesting question how these two aspects of the dataset interact \cite{Luo_etal2019}.

After having tested our assumptions on the simulations above, we will extend our observations (above and those on page \pageref{properties1}) on generalized penalty functions to real datasets.
\FloatBarrier

\section{\label{subsec:distance threshold choice}Choosing the distance threshold for circular coordinates computation}
In this appendix, we discuss the effect of the choice of \emph{distance threshold $r$} in the smoothing at step 4 of the algorithm to compute circular coordinates. The threshold identifies the Vietoris-Rips complex on which to compute the smoothing of the cocycle.
This complex is built on the points at a mutual distance at most $r$ from other points.

\begin{figure}[t!]
\begin{adjustbox}{center}

\includegraphics[width=1.2\textwidth]{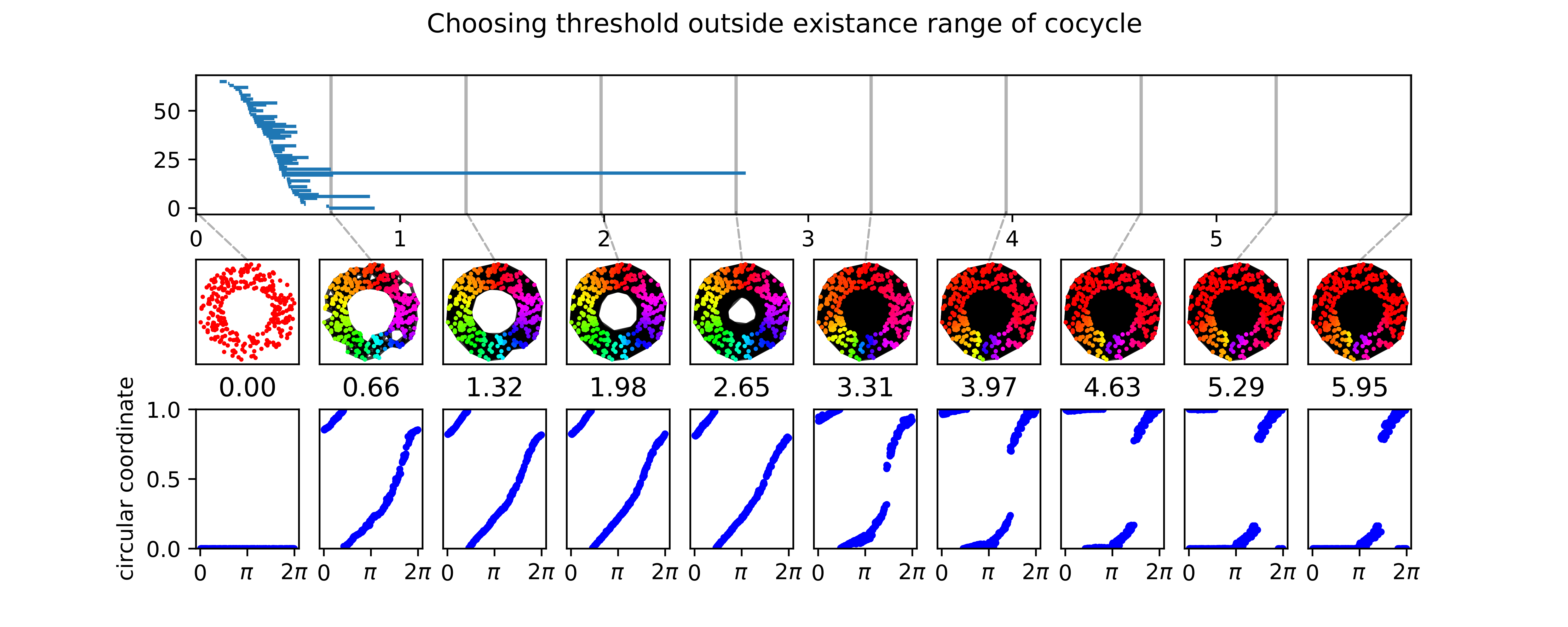}
\end{adjustbox}
\caption{\label{fig:threshold_not} (top) Barcode for a simulated example of 150 uniformly sampled points from an annulus. (bottom) Resulting circular coordinates computed using different thresholds along the filtration for longest persisting cocycle represented as color of the points.}
\end{figure}

As long as the distance threshold $r$ is within the range of birth and death for the selected persistent cocycle (Fig. \ref{fig:threshold_not}), the choice of Vietoris-Rips filtration threshold does not influence the outcome of the computation of the circular coordinates.

\begin{figure}
\begin{adjustbox}{center}
\includegraphics[width=1.2\textwidth]{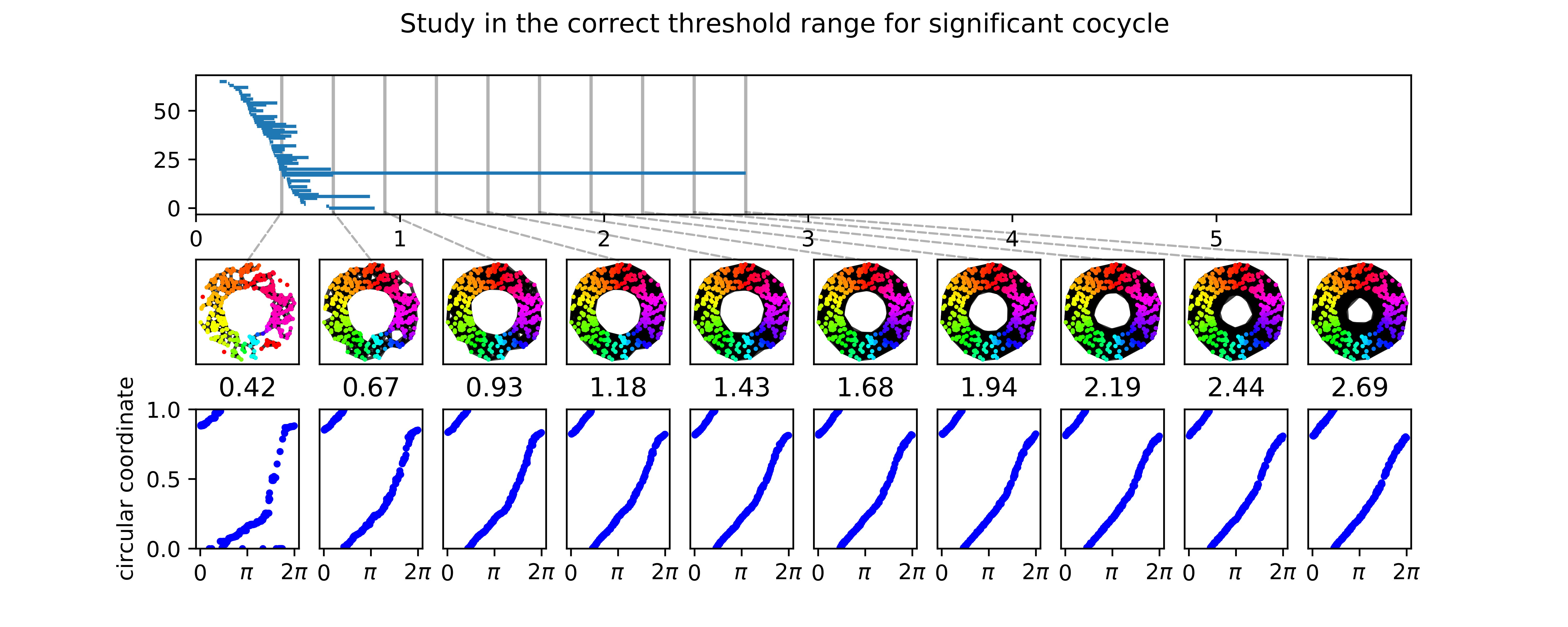}
\end{adjustbox}
\caption{\label{fig:threshold_yes} (top) Barcode for a simulated example of 150 uniformly sampled points from an annulus. (center) Resulting circular coordinates computed using different thresholds along the filtration for longest persisting cocycle represented as color of the points. (bottom) Resulting circular coordinates plotted against the angle theta between the respective point and the $x=0$ axis with values colored the same way as the center row.}
\end{figure}

Any choice of a distance threshold within this range will induce a comparable circular coordinate in the connected component containing the longest surviving cocycle. In Fig. \ref{fig:threshold_yes} we show how for values close to the birth of the cocycle, the relative simplicial complex has more than one connected component. In this case, the circular coordinate is computed only in the component containing the selected cocycle. Furthermore, comparing the 100 circular coordinates computed with the threshold varying between the birth and death of the longest cocycle in the example Fig. \ref{fig:threshold_yes}, we can see that the values of the coordinates are the same up to a small $\epsilon$. The only exceptions are the values corresponding to the small component that for small values of the threshold does not contain the cocycle and for larger values merges with the giant connected component.

\begin{figure}
\begin{adjustbox}{center}
\includegraphics[width=1\textwidth]{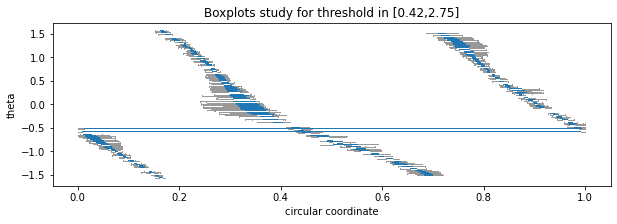}
\end{adjustbox}
\caption{\label{fig:threshold_mean} Comparison of 100 circular coordinates computed with threshold varying between the birth and death of the longest cocycle in the example Fig. \ref{fig:threshold_yes}. The blue bars represent a box plot for the circular coordinate values for the circular coordinates relative to the points represented by angle theta.}
\end{figure}

\FloatBarrier

\medskip
Received March 2021; revised June 2021; early access September 2021.
\medskip

\end{document}